\documentclass[default,iicol]{sn-jnl}

\usepackage{algorithm}
\usepackage{algpseudocode}  
\usepackage{multirow}
\usepackage{booktabs}
\usepackage{subfigure}
\usepackage{hyperref}
\usepackage{color,soul}

\jyear{2021}%

\theoremstyle{thmstyleone}%
\theoremstyle{thmstyletwo}%

\theoremstyle{thmstylethree}%

\begin{document}

\title[Article Title]{Learning Graph Representation of Person-specific Cognitive Processes from Audio-visual Behaviours for Automatic Personality Recognition}

\author[1]{\fnm{Siyang} \sur{Song}}\email{ss2796@cam.ac.uk}

\author[2,3,4]{\fnm{Zilong} \sur{Shao}}\email{shaozilong2019@email.szu.edu.cn}

\author[5]{\fnm{Shashank} \sur{Jaiswal}}\email{shashank@blueskeye.com}

\author*[2,3,4]{\fnm{Linlin} \sur{Shen}}\email{llshen@szu.edu.cn}

\author[5]{\fnm{Michel} \sur{Valstar}}\email{Michel.Valstar@nottingham.ac.uk}

\author[1]{\fnm{Hatice} \sur{Gunes}}\email{Hatice.Gunes@cl.cam.ac.uk}
\affil[1]{\orgdiv{Department of Computer Science and Technology}, \orgname{University of Cambridge}, \orgaddress{\city{Cambridge}, \country{United Kingdom}}}

\affil[2]{\orgdiv{Computer Vision Institute}, \orgname{Shenzhen University}, \orgaddress{\city{Shenzhen},  \country{China}}}

\affil[3]{\orgname{Shenzhen Institute of Artificial Intelligence of Robotics of Society}, \orgaddress{\city{Shenzhen},  \country{China}}}

\affil[4]{\orgdiv{Guangdong Key Laboratory of Intelligent Information Processing}, \orgname{Shenzhen University}, \orgaddress{\city{Shenzhen},  \country{China}}}

\affil[5]{\orgdiv{Computer Vision Lab}, \orgname{University of Nottingham}, \orgaddress{\city{Nottingham},  \country{United Kingdom}}}


\abstract{This paper proposes to recognise the true (self-reported) personality from the learned simulation of the target subject's cognition. This approach builds on two following findings in cognitive science: (i) human cognition partially determines expressed behaviour  and is directly linked to true personality traits; and (ii) in dyadic interactions individuals’ nonverbal behaviours are influenced by their conversational partner's behaviours. In this context, we hypothesise that during a dyadic interaction, a target subject's facial reactions are driven by two main factors, i.e. their internal (person-specific) cognitive process, and the externalised nonverbal behaviours of their conversational partner. Consequently, we propose to represent the target subject’s (defined as the listener) person-specific cognition in the form of a person-specific CNN architecture that has unique architectural parameters and depth, which takes audio-visual non-verbal cues displayed by the conversational partner (defined as the speaker) as input, and is able to reproduce the target subject’s facial reactions. Each person-specific CNN is explored by the Neural Architecture Search (NAS) and a novel adaptive loss function, which is then represented as a graph representation for recognising the target subject’s true personality. Experimental results not only show that the produced graph representations are well associated with target subjects' personality traits in both human-human and human-machine interaction scenarios, and outperform the existing approaches with significant advantages, but also demonstrate that the proposed novel strategies such as adaptive loss, and the end-to-end vertices/edges feature learning, help the proposed approach in learning more reliable personality representations. Building on our earlier version of this work, this paper further proposes: (i) assigning a unique depth for each CNN; (ii) a novel end-to-end graph vertex feature learning strategy; (iii) a transformer-based edge feature learning strategy; and (iv) evaluating the approach in human-machine interaction scenario.}




\maketitle

\section{Introduction}
\label{sec:intro}

\noindent Personality is defined as the characteristic set of human external verbal and non-verbal behaviours, as well as internal cognition and emotional patterns that evolve from biological and environmental factors \cite{hogan1997handbook}, which partially reflect the subject's identity. Understanding human personality can benefit a wide range of applications such as (mental) health condition analysis \cite{jaiswal2019automatic,lo2017genome}, candidate screening for recruitment \cite{PonceEtAl-ECCV16}, as well as personalised, adaptive human-agent interactions (e.g., \cite{GunesEtAl-2019}).

Recent advances in machine learning (ML) have enabled the development of non-invasive automatic personality traits analysers that recognise subjects' personality traits from their audio-visual non-verbal behaviours \cite{li2020cr,zhang2019persemon,ventura2017interpreting,fang2016personality,celiktutan2017automatic,song2021self} as there is solid psychological and biological evidence \cite{eysenck2005cognitive,Willerman1994brain,corrigall2013music,kumari2004personality} claiming that nonverbal behaviours are reliable predictors of personality. In most of these approaches, ML models are trained with the personality labels provided by the external observers (annotators), and they therefore output their \emph{perception} of the target subjects' personality. In other words, these ML models play the role of an external artificial observer that observes the target subjects' nonverbal distal cues \-- i.e., audio signals (e.g., delta-mel-cepstral, speech duration, pitch, and pause rate, etc.) \cite{fang2016personality,celiktutan2015automatic,liu2020speech,an2018lexical}, visual cues (e.g., facial actions and gestures) \cite{principi2019effect,wei2018deep,guccluturk2016deep}, observable inter-personal cues \cite{salam2016fully,fang2016personality,CeliktutanEtAl-TAC2017} etc., and output the external observer's perception of target subjects' personality. People externalize their personality through distal cues (e.g., energy) but these cues undergo a perception bias based on what the observer actually perceives, and become proximal cues (e.g., loudness). As a result, these approaches can be treated as Automatic Personality Perception (APP) solutions (inference from proximal cues) \cite{vinciarelli2014survey}.

\begin{figure*}
	\centering
	\subfigure[Searching a person-specific processor to simulate a target subject's (listener's) cognition.]{\label{subfig:P-CNN}
		\includegraphics[width=16cm]{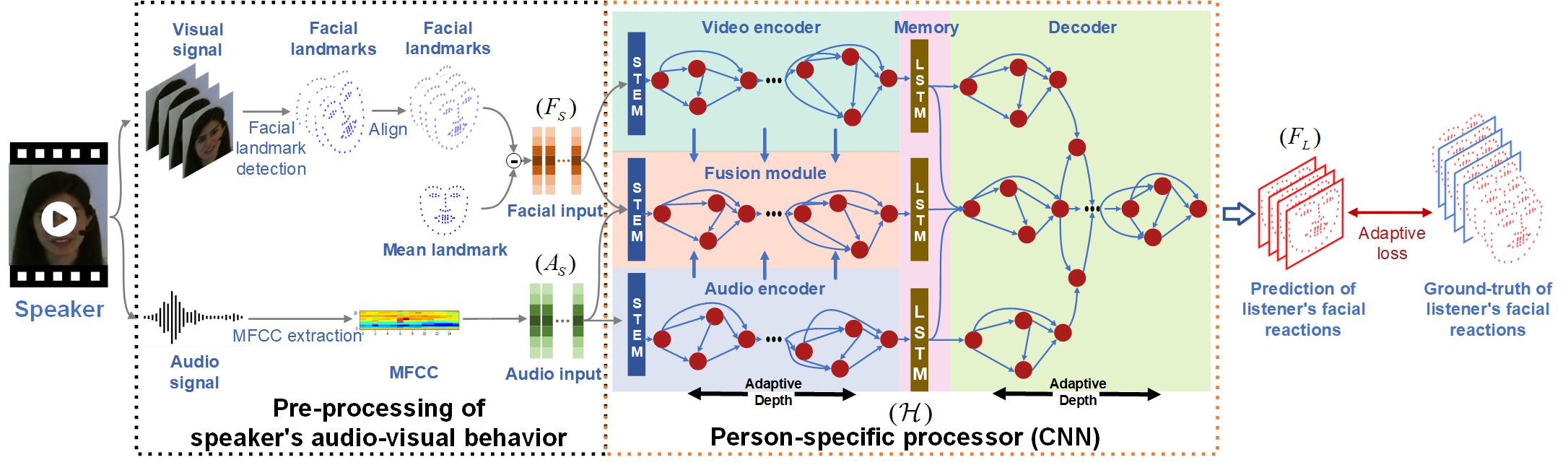}}
	\subfigure[Recognition of the target subject's (listener) personality from the graph representation of the target subject's (listener) person-specific CNN model.]{\label{subfig:Graph}
		\includegraphics[width=16cm]{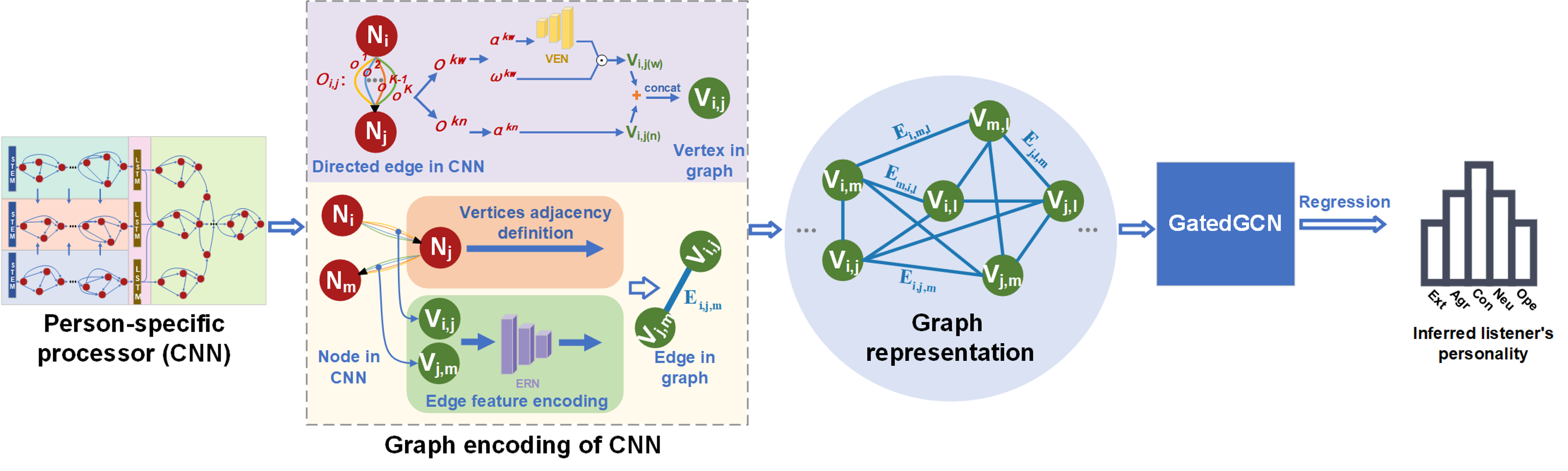}}
	\caption{The pipeline of the proposed approach. (a) Our approach starts with searching a person-specific processor (multi-modal CNN) architecture (unique topology, weights and depth) that can reproduce listener's facial reactions according to the speaker's audio-visual non-verbal signals (Sec. \ref{subsec: simulate cognition}); (b) Then, we parameterize the person-specific processor as a graph representation to represent the listener's cognition and feed it to a graph neural network for listener's personality recognition (Sec. \ref{subsec: graph representation}).} 
    \label{fig:pipeline}
\end{figure*}

In some scenarios, the goal is to infer true personality from machine detectable distal cues, i.e., Automatic Personality Recognition (APR) \cite{vinciarelli2014survey}. While APP approaches predict apparent personality (perception) based on proximal behavioural cues, APR aims to recognise the true personality that impacts the generation of distal behavioural cues. Thus, APP models that were trained as external observers to provide personality perceptions may not be reliable for recognising true personality traits (\textbf{Problem 1}). Moreover, the majority of these APP solutions  \cite{wei2018deep,fang2016personality,kaya2017multi,guccluturk2016deep,curto2021dyadformer} recognise personality traits from every single frame or thin slice of behaviour, independently, by re-using clip-level personality labels as the frame/thin slice-level labels to train ML models that can provide a personality prediction for each frame/thin slice. This is problematic as people with different personality traits may express very similar non-verbal audio-visual behaviours in a single frame or a thin slice. As a result, such strategies may lead to the same input pattern being paired with multiple labels during training, making it practically impossible to learn a good hypothesis \textbf{(Problem 2)}. Although some recently proposed approaches \cite{li2020cr,zhang2019persemon,beyan2019personality} address this issue by modelling personality using entire clip, (i.e., recognising personality traits at the clip-level), they only select a set of key frames to represent an entire clip. This results in ignoring the short-term behaviours displayed by the discarded frames (\textbf{Problem 3}), which may contain crucial cues for personality recognition.

In this paper, we propose a novel audio-visual automatic personality recognition framework that addresses the problems highlighted above. The proposed approach aims to infer true personality, as it is built on the definition that personality influences the individual's externalization of their distal cues \cite{vinciarelli2014survey} (e.g., the cognitive processes for facial reaction generation). This is achieved by learning a person-specific CNN that can represent the target subject's cognitive process for facial reaction generation and then encoding the architecture of the explored CNN as a graph representation for personality recognition (\textbf{addressing Problem 1}). Since CNNs can accurately simulate various human cognitive processes (e.g., pattern detection, classification, regression, reconstruction, etc), we hypothesize that if a CNN can accurately represent the target subject's cognitive processes for facial reaction generation under various situations (e.g., different conversations), the cognitive processes of this CNN (decided by the CNN's architecture and weights) are also well associated with the target subject's true personality. Recent works \cite{Busso-TAC2013,MouEtAl-2019} show that in dyadic and group interactions subjects’ nonverbal behaviours are influenced by, and therefore can be predicted from, the behaviours of their conversational partner(s). Therefore, this paper assumes that during a dyadic interaction, the target subjects' (listeners) facial reactions are driven by two main factors: (i) the target subject's internal (person-specific) cognition, and (ii) the externalised nonverbal behaviours of the conversational partner (the speaker). Consequently, we propose to learn a person-specific CNN for reproducing the target subject's facial reaction in response to the conversational partner. More importantly, each person-specific CNN is learned using all available frames/thin slices of the target video without ignoring any frames (\textbf{addressing Problem 3}) and thus the explored person-specific CNN architecture contains the clip-level information of the target subject, which is then encoded as a graph representation. As a result, the training of the personality recognition model is implemented by pairing the clip-level graph representation with the clip-level personality labels, \textbf{avoiding Problem 2}. The full pipeline of the proposed approach is illustrated in Fig. \ref{fig:pipeline}. 

The main contributions and novelty of this paper can be summarised as follows:
\begin{itemize}

    \item  To the best of our knowledge, this is the first audio-visual approach that attempts to recognise the true (self-reported) personality traits of target subjects by modelling their cognitive processes.
    
    \item We propose a novel audio-visual non-invasive human person-specific cognition simulation strategy which automatically searches for an optimised multi-modal person-specific CNN architecture to reproduce the target subject's facial reactions, where a novel adaptive loss function is proposed to model the uncertain time delay of the human facial reactions. The explored person-specific CNN has a unique combination of layers (operations), weights and depth, which plays the role of the target subject's cognitive process for facial reaction generation.

    \item We propose novel graph encoding strategies to parameterize a CNN with an arbitrary architecture and depth either into an architecture and depth-independent graph representation or a heterogeneous graph representation, where each directed edge (contains a set of operations) in the CNN is treated as a vertex in the graph representation and the adjacency of vertices are decided by whether their corresponding directed edges are connected to the same node in the CNN. In particular, an end-to-end learning strategy is proposed to encode CNN's operation parameters (OPs) and layers' weights (LWs) of each directed edge as an vertex feature from the perspective of the target (personality recognition). A multi-modal transformer is also proposed to learn a multi-dimensional feature for each edge of the graph representation, which models the detailed relationship between adjacent vertices. Both strategies aim to deep learn task-specific information from the explored CNN to provide richer and reliable cues for personality recognition. To the best of our knowledge, this is the first audio-visual approach that combines the transformer and graph neural networks (GNNs) for personality recognition.

    \item We conduct a set of experiments under both human-human and human-machine dyadic interaction settings, which not only validate the superior performance of the proposed approach in recognising true personality traits but also systematically demonstrate the influence of the various internal (methodological components) and external (subject demographic groups) factors on the proposed approach in recognising true personality traits.

\end{itemize}

An earlier version of this work was presented in \cite{shao2021personality}, but it has several limitations: (i) while depth is a key factor that impacts the cognitive processes of CNNs, the previous work was setting all person-specific CNNs to have the same depths; (ii) the vertices feature was obtained by simply multiplying the OPs with the corresponding LWs without taking the target (personality recognition) into consideration; (iii) it only used a simple 1D CNN for edge feature learning without specifically modelling the relationship between corresponding vertices; (iv) the approach was only evaluated on a human-human dyadic interaction dataset; and (v) didn't explain the implementation details of the reproduced approaches.  This paper extends our previous work to address these limitations, and makes the following additional contributions:
\begin{itemize}

    \item \textbf{Methodological contributions:} Firstly, we introduce a depth search strategy in Sec. \ref{subsec: multimodal}, allowing each person-specific CNN to not only have unique architectural parameters but also a unique depth. In addition to align all person-specific CNNs as graph representations with the same size, we further propose to encode CNNs of variable depths as heterogeneous graph representations. Secondly, we propose a novel end-to-end vertex feature learning strategy to encode task-specific vertex features from corresponding OPs and LWs (Sec. \ref{subsec: vertex feature}). Finally, we propose a novel transformer-based edge feature learning strategy that employs attention operations to learn salient task-specific relationship between vertices (Sec. \ref{subsec:edge_feat});

    \item \textbf{Scientific contributions:} We evaluate the proposed approach under both human-human and human-machine dyadic interaction and systematically investigate the influence of the various subject demographic and methodological factors on the proposed approach. We also present the implementation details of the reproduced approaches in Sec. \ref{subsec:imple}.
    
\end{itemize}


\section{Related Work}
\label{sec: Related}

\noindent In this section, we first review previous audio-visual automatic personality analysis approaches, most of which are APP solutions that directly predict apparent personality from target subjects' observable audio-visual behaviours (Sec.\ref{subsec:rela-automatic}). We then summarize biological and psychological studies which found that personality can be reflected by human cognition, providing the theoretical basis for our work, i.e., recognising true personality traits from the simulated human cognitive processes (Sec.\ref{subsec:related-relationship}). Since this paper extends the neural architecture search (NAS) technique to search for person-specific CNN architectures, and uses GNNs for personality recognition, we also briefly review the previous NAS and GNN approaches in Sec. \ref{subsec:rela-NAS} and Sec. \ref{subsec:rela-graph}, respectively.



\subsection{Audio-visual automatic personality analysis}
\label{subsec:rela-automatic}

\noindent Early audio-visual automatic personality analysis approaches usually extract hand-crafted features to describe audio-visual human non-verbal behaviours or interpersonal relationship between subjects, including low-level features such as histogram of oriented gradients (HOG) \cite{joshi2014automatic},  Local Phase Quantization (LPQ) \cite{eddine2017personality}, etc., and mid-level cues such as statistics of mid-level behaviour attributes \cite{teijeiro2015your,fang2016personality}, facial attributes (gazes, head motions, etc) \cite{lepri2012connecting}, human postures and gestures cues during the speaking \cite{nguyen2013multimodal}, co-occurrent patterns of behaviours \cite{okada2015personality}, body skeleton activity \cite{salam2016fully}, Quantised Local Zernike Moments (QLZM) \cite{celiktutan2017automatic}, visual focus of attention \cite{salam2016fully}, etc. These hand-crafted features are then fed to traditional machine learning models such as support vector machine regressor (SVR) or logistic regression to generate apparent personality predictions.

Due to recent advances in deep learning, most recently proposed approaches employ Convolution Neural Networks (CNNs) to learn task-specific deep features from each frame or a thin video slice. For example, Ventura et al. \cite{ventura2017interpreting} propose a Descriptor Aggregation Network (DAN) to extract a frame-level feature at multiple spatial resolutions, and use such multi-level visual features to infer personality at the frame-level. To learn both personality-related audio and visual cues, two-stream bi-modal networks are proposed in \cite{guccluturk2016deep} and \cite{wei2018deep}, which firstly learn frame-level audio-visual features and then combine them at the the fully connected layer to provide frame-level personality prediction. The video-level prediction is then made by averaging predictions of all frames. Principi et al.\cite{principi2019effect} propose a multi-modal CNN to jointly learn audio and visual information from every image sequence (thin video slice) and audio segment. The extracted features are combined with attribute-specific models to predict personality traits.

Since personality trait models focus on evaluating the aspects of personality that are relatively stable over a long period of time for the target subject \cite{kassin2003essentials} (usually be much longer than the duration of a single audio-visual clip), the frame/thin slice-level behaviours may not be reliable in reflecting personality traits \cite{kassin2003essentials}. Consequently, approaches that model personality traits based on clip-level/long-term behaviours also have been frequently investigated. One popular solution is to summarise frame/thin slice-level features of an entire audio-visual clip into a global statistical descriptor to infer personality \cite{bekhouche2017personality,nguyen2013multimodal}, e.g., averaging all frame/thin slice-level vectors or using histogram to represent frame/thin slice-level feature distributions. To consider important dynamic cues, Subramaniam et al. \cite{subramaniam2016bi} employ Long-short-term-memory Networks (LSTMs) to encode the short-term and long-term temporal information into  frame-level deep-learned features and the video-level prediction is made by the decision-level fusion of frame-level predictions. Besides, another popular way to learning clip-level representation is selecting a set of key frames from an entire video. Zhang et al. \cite{zhang2019persemon} selects a face image from every thin slice to represent the given video. Then, a consensus strategy is employed to process all the selected frames, to produce video-level personality predictions. Similarly, Li et al. \cite{li2020cr} first divide the video into $32$ slices, and then randomly select a face image and a face-background image from each of them, which are then stacked as the clip-level stream. This approach also converts the acoustic wave of an entire clip to fixed-length vectors as the clip-level audio representation. Subsequently, the video-level prediction is made by these selected frames. Beyan et al. \cite{beyan2019personality} propose to generate multiple dynamic facial images \cite{bilen2016dynamic,song2019dynamic,song2021dynamic} to represent each video segment and then choose a set of dynamic facial images that have the highest spatio-temporal saliency as the key frames to construct the video-level representation.

In summary, while modelling personality traits at the frame/segment-level is problematic, the recent clip-level representations usually failed to utilise the full scale of the available information in the data, as they select a subset or key frames to represent an entire video. To avoid these problems, Song et al. \cite{song2021self} propose a domain adaption approach to learn a set of intermediate convolution layers from all available data as the person-specific representation for the target subject, which achieved a comparable performance to the state-of-the-art method \cite{li2020cr}. However, similar to the approaches described above, it still directly infers apparent personality based on the subjects' observable behaviours. In other words, all the aforementioned studies are proposed to predict personality perception.

\subsection{The relationship between personality and human cognition}
\label{subsec:related-relationship}

\begin{figure}
\centering
\includegraphics[width=8cm]{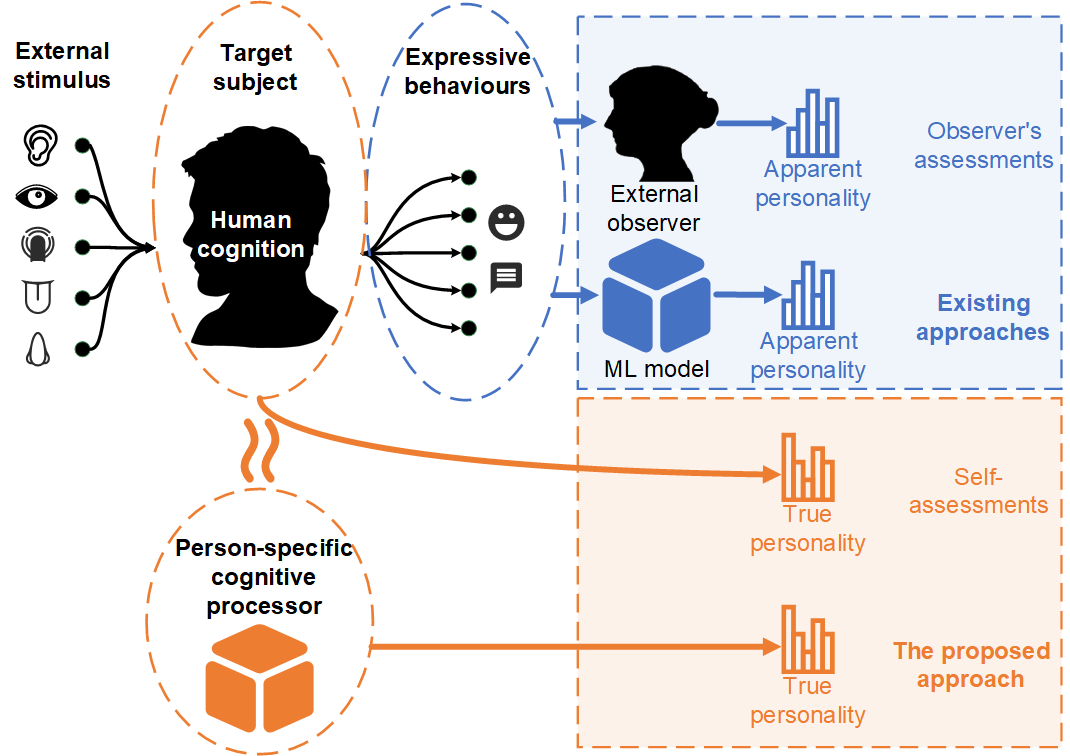}
\caption{The difference between the proposed approach (depicted in orange) and existing approaches (depicted in blue). While the existing approaches attempt to achieve automatic personality perception from non-verbal behaviours, our approach learns to recognise true personality by modelling target subjects' cognition.}
\label{fig:comparsion}
\end{figure}

\noindent According to previous biological studies \cite{eysenck2005cognitive,Willerman1994brain}, personality traits (e.g., Extraversion, Conscientiousness, and Neuroticism) are well associated with human brain structures \cite{kernberg2016personality} and activities such as brain local volumes \cite{deyoung2010testing} and gray and white matter \cite{jackson2011exploring}, which are key factors in deciding and controlling human cognitive processes. For example, Kumari et al. \cite{kumari2004personality} investigated brain fMRI activity based on “n-back” task, and found that brain responses during cognitive activities are related to extraversion and neuroticism traits. Previous psychological studies also frequently claimed that people's personality is well associated with their cognitive processes in various daily activities such as risk taking \cite{kogan1964risk}, creativity \cite{mccrae1993openness,corrigall2013music}, music learning \cite{corrigall2013music}.  An exploratory factor analysis was conducted by \cite{corrigall2013music}, and the results show that creativity and primary cognitive processes are correlated with the extraversion and psychoticism (neuroticism) traits. Importantly, the relationship between human cognition and personality are relatively stable, as a longitudinal study conducted by Schaie et al. \cite{schaie2004seattle} showed that some of the personality-cognition relations could last for over 35 years. This finding gives us the inspiration that human cognition can be a reliable and stable source for recognising personality. As reviewed in Sec. \ref{subsec:rela-automatic}, the main difference between our approach and the existing approaches (illustrated in Fig.~\ref{fig:comparsion}), is the fact that the existing approaches attempt to achieve automatic personality perception directly from observable non-verbal behaviours of the target subjects, where the ML model acts as an external observer. Instead, our approach draws inspiration from the aforementioned works on the interrelationship between personality and human cognition, and learns to recognise true personality by simulating and modelling target subjects' person-specific cognitive processes.

\subsection{Neural architecture search}
\label{subsec:rela-NAS}

\noindent We also briefly review previous Neural Architecture Search (NAS) studies as we extend this technique to design our person-specific CNNs. NAS is an Auto-ML technique that can replace humans to automatically design neural network architectures for many machine learning tasks. Most early NAS approaches were based on evolution algorithms \cite{real2017large,xie2017genetic} or reinforcement learning \cite{baker2016designing,zoph2016neural}, which use discrete strategy to search for optimal network architectures. Consequently, such methods usually suffer from extended period of searching time, e.g., the approach of \cite{zoph2016neural} requires $22,400$ GPU days to search for an optimal network on the CIFAR-10 dataset. The recent advances in modular-based approaches \cite{zoph2018learning,pham2018efficient} provide faster solutions. They construct an entire network by repeatedly stacking a set of 'modules' which can be a pre-defined unit operation or a small cell that contains multiple operations. In particular, Pham et al. \cite{pham2018efficient} proposed to construct a large network with multiple sub-networks (cells) that share parameters. This strategy can largely accelerate the searching speed while still allowing the explored network to have strong generalization capability. As a result, it has been widely extended in recent NAS applications \cite{zhong2018practical,dong2018dpp,liu2018darts}.

To further reduce the searching duration, some studies \cite{shin2018differentiable,liu2018darts,luo2018neural} not only employ the cell-based parameter sharing strategy, but also propose continuously differentiable algorithms to replace discrete searching strategy, which allows the use of gradient optimization techniques for faster adjustment of the architectural parameters. The DARTs algorithm \cite{liu2018darts} assumes that a cell is a directed acyclic graph (DAG) consisting of an ordered sequence of nodes, where each node contains a set of feature maps. These nodes are connected by directed edges, each of which contains a set of pre-defined operations (e.g., convolution, pooling, etc.). Then, the entire network can be constructed by stacking multiple cells that share the same set of operation parameters (OPs), i.e., the importance of each operation. During the network searching, the OPs and layers' weights (LWs) of these operations are jointly optimized based on a bi-level optimization strategy which separately optimizes OPs in the validation set and LWs in the training set. This continuously differentiable NAS algorithm allows the searching time in CIFAR-10 to be reduced to $1$ GPU day, while still retaining the best classification performance. 

\subsection{Graph Neural Networks}
\label{subsec:rela-graph}

\noindent Recently, there is increasing interest in utilising the advantages of deep learning to develop Graph Neural Networks (GNNs) and Graph Convolution Neural Networks (GCNs) \cite{wu2020comprehensive}. Among various types of GNNs and GCNs, the message-passing GNNs have shown superior capabilities for leveraging the standard deep learning operations such as batching, residual connections and normalization, which achieved superior performance compared to other graph models across various applications \cite{dwivedi2020benchmarking}. Although the standard GNN \cite{scarselli2008graph} and GCN \cite{defferrard2016convolutional} have already shown strong capabilities in graph classification/regression tasks, the recently proposed Graph Attention Network \cite{velivckovic2017graph} that uses masked self-attentional layers and Residual Gated Graph Neural Network (Gated-GCN) \cite{bresson2017residual} with residual connection and gate mechanism have achieved the state-of-the-art performance on many graph-based tasks \cite{dwivedi2020benchmarking}. As a result, this paper employs the Gated-GCN as the personality recognition model.

When constructing the graph representation for raw data, the majority of existing approaches \cite{scarselli2008graph,defferrard2016convolutional,velivckovic2017graph,bresson2017residual} used a binary value (0 or 1) or a single-dimension weight \cite{isufi2020edgenets,wang2021graphtcn} as the edge feature to define the connectivity or a simple relationship between a pair of vertices. To provide richer relationship information, Gong et al. \cite{gong2019exploiting} successfully explored a way to use multi-dimensional hand-crafted feature as the edge feature for GNNs-based task. Inspired by its superiority, which also has been validated by several recent works \cite{mavroudi2020representation,lei2020novel,dwivedi2020benchmarking}, this paper proposes the first approach to deep learn task-specific multi-dimensional edge features in an end-to-end manner for personality recognition.

\section{Methodology}
\label{sec: method}

\noindent In this section, we present a novel approach which models person-specific cognitive processes from target subjects' audio-visual non-verbal behaviours, aiming to recognise true personality that is associated with human expressive distal cues (e.g., facial actions) governed by human cognition.

\textbf{Overview:} Our approach starts with simulating and modelling the target subject's cognitive processes. Specifically, we search for an optimal person-specific multi-modal CNN architecture for each target subject, which can accurately reproduce the subject's facial reactions according to the external audio-visual signals that are received by the subject (explained in Sec. \ref{subsec: simulate cognition} and Fig. \ref{subfig:P-CNN}). As a result, we hypothesize that the explored person-specific CNN architecture can represent the person-specific cognitive processes of the target subject. To use the explored CNN as the target subject's person-specific representation for personality recognition, we parameterize its operation parameters (OPs), layers' weights (LWs) and depths into a fixed-size graph representation which is then processed by a GatedGCN model for personality recognition (explained in Sec. \ref{subsec: graph representation} and Fig. \ref{subfig:Graph}).

\textbf{Novelty:} The main novelties of the proposed approach are summarised as follows: firstly. we propose to use the simulated human cognition as the source descriptor to recognise true personality traits, which differs from existing approaches \cite{song2021self,li2020cr,zhang2019persemon,beyan2019personality,celiktutan2017automatic,guccluturk2016deep,principi2019effect,wei2018deep} that predict apparent personality traits directly from target subjects' expressive behaviours. Secondly, we propose the first non-invasive approach that simulates human person-specific cognitive processes that relate to facial reactions. Thirdly, we propose a novel graph representation that encodes the architectural parameters of a person-specific CNN as the person-specific descriptor. This differs from all existing deep learning-based approaches \cite{li2020cr,zhang2019persemon,beyan2019personality} that use the latent features of the input image/video as personality descriptor. Fourthly, the proposed graph representation can encode an arbitrary number of architectural parameters of a person-specific CNN into a length-independent video-level descriptor, which is learned from the target subject's behaviours of the entire video. This strategy neither models the personality at the frame/thin slice-level \cite{celiktutan2017automatic,wei2018deep} nor discards any frames \cite{li2020cr,zhang2019persemon}. Fifthly, in comparison to most GNN-based studies that only consider the binary adjacency relationship between a pair vertices \cite{scarselli2008graph,defferrard2016convolutional,velivckovic2017graph,bresson2017residual}, or use hand-crafted descriptors \cite{gong2019exploiting} to model the relationship between vertices in a graph, we propose to deep learn task-specific multi-dimensional edge features in an end-to-end manner. In comparison to our earlier version \cite{shao2021personality}, this paper: (i) allows each person-specific CNN to have a unique depth; (ii) replaces the hand-crafted vertices feature encoding with end-to-end deep vertex feature learning; (iii) replaces the simple 1D CNN with a transformer-based network for end-to-end edge feature learning; and (iv) further proposes two topology alignment strategies to encode CNNs of variable-depths into fixed-size graph representations.

\subsection{Simulating person-specific cognition}
\label{subsec: simulate cognition}

\noindent This section explains how we search for a person-specific CNN architecture to simulate target subject's cognitive process. The explored person-specific CNN is expected to have a unique architecture, depth and weights. Therefore, we explicitly describe each part of our approach, including the input and target (Sec. \ref{subsec: input_target}), basic CNN topology to represent human cognitive processes (Sec. \ref{subsec: multimodal}), loss function for CNN's searching and training (Sec. \ref{subsec: adaptive loss}), and the optimization strategy (Sec. \ref{subsec: optimization}).

\subsubsection{Input and target}  
\label{subsec: input_target}

\begin{figure}
\centering
\includegraphics[width=8cm]{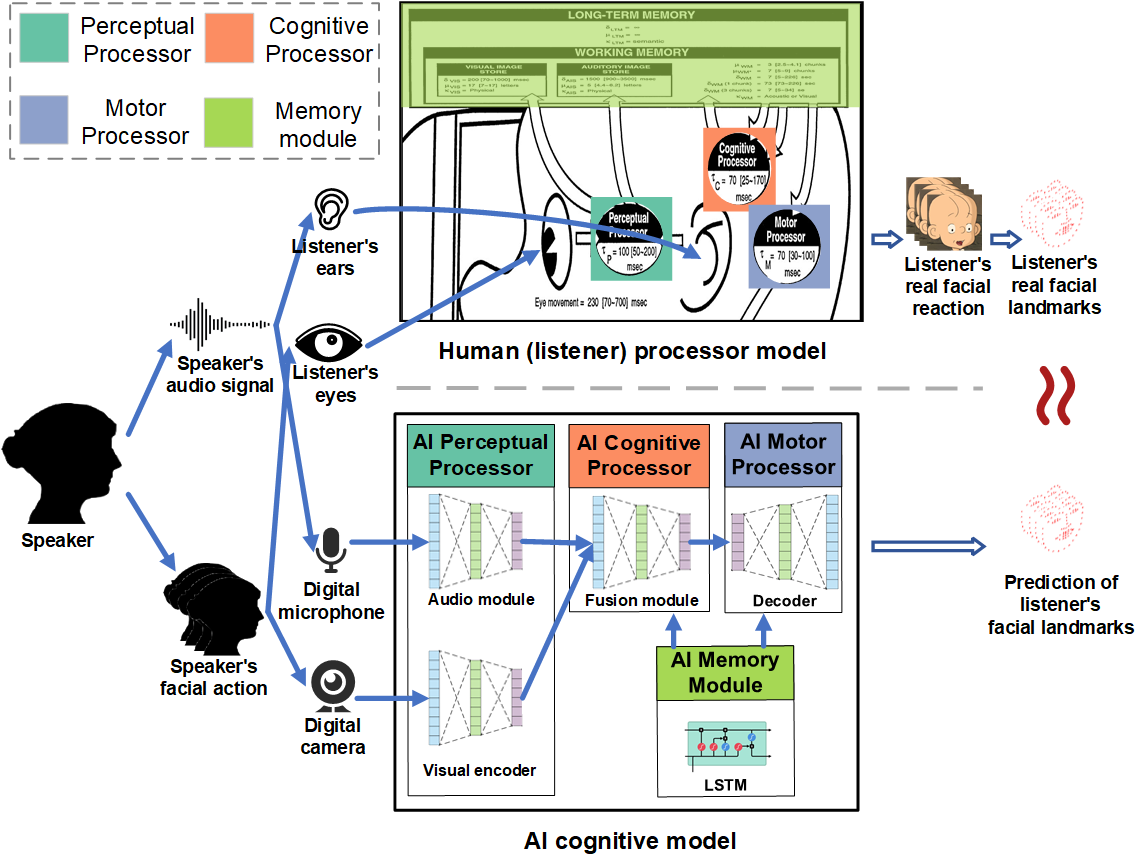}
\caption{Using CNN to simulate human cognition.}
\label{fig:hypothesis}
\end{figure}

\noindent We first define the criteria for evaluating whether a CNN is able to simulate an subject's cognitive process accurately. Inspired by the Model Human Processor (MHP) \cite{card1986model} (shown in Fig. \ref{fig:hypothesis}), which receives and processes external signals and accordingly generates behavioural reactions, we hypothesise that when a CNN receives the same external signals as the target subject, if it is able to generate the same reactions as the target subject, its architecture can represent the subject's person-specific cognitive process. 

Let's recall our hypothesis: during a dyadic interaction, the listener's facial reactions are driven by two main factors: (i) listener's internal (person-specific) cognition, and  (ii) the externalised nonverbal behaviours of the conversational partner (referred to the \textbf{speaker} in this paper). Then, we formulate our problem as follows. During a dyadic interaction, given an audio signal $A_{S}$ and facial behaviours $F_{S}$ of the speaker, the goal is to learn a hypothesis (a CNN model) $\mathcal{H}$ that can reproduce the facial reactions $F_{L}$ of the listener, which can be formulated as:
\begin{equation}\label{hypo}
F_{L} = \mathcal{H}(A_{S}, F_{S})
\end{equation}
Once $\mathcal{H}$ is well learned, it takes on the role of the corresponding listener's cognitive processor in generating facial reactions during a dyadic interaction. Consequently, we assume that the learnt $\mathcal{H}$ is \emph{sufficiently} informative for modeling the listener's true personality traits not only because individuals' true personality relate to their cognitive processes but also true personality is a key factor in governing how human non-verbal behaviours are generated and displayed \cite{vinciarelli2014survey}.

In this paper, we use speaker's and listener's facial landmark sequences (in the form of $80$ frames) to represent the input and target facial movements, respectively. The aligned facial landmarks are obtained for each frame using OpenFace 2.0 \cite{baltrusaitis2018openface}, which are then transformed based on a pre-defined mean face shape in order to keep only facial behaviours without the identity information (as suggested by \cite{eskimez2018generating}). Also, we use $64$ bin log-mel spectra as the audio representation, where each audio frame is computed by a $40$ ms hanning window with stride size of $40$ ms. This way, the number of audio frames for each video is the same as the number of video frames. This process is illustrated as the pre-processing part in Fig. \ref{subfig:P-CNN}.

\subsubsection{Multi-modal cognitive processor model} 
\label{subsec: multimodal}

\noindent Let's recall that the main idea of our approach is to recognise true personality traits from the simulated human cognitive process. To this end, we represent the person-specific cognitive process $\mathcal{H}$ of the target subject as a person-specific multi-modal CNN whose architecture is automatically searched. Inspired by the Model Human Processor \cite{card1986model}, the main topology of each person-specific CNN is set to have a visual encoder and an audio encoder that simulate the \textit{human perceptual processor}, an audio-visual decoder that simulates the \textit{human motor processor} and a fusion module that partially simulates the \textit{human cognitive processor}, i.e., jointly processing audio-visual cues at multiple levels. In addition, a Long-Short-Term-Memory network (LSTM) is employed to process the latent features generated from two encoders and the fusion module, simulating the \textit{human working memory} (illustrated in Fig. \ref{fig:hypothesis}). In this paper, we set the audio encoder, the visual encoder and the fusion module to have the same number of blocks, including several regular blocks and down-sampling blocks. Specifically, each regular block is made up of multiple stacked cells, where each cell is made up of a set of nodes and directed edges (explained below), and the input and output feature maps of each regular block have the same size.

\begin{figure}
	\centering
	\subfigure[Illustration of the fusion module details.]{\label{subfig:fusion_module-CNN}
		\includegraphics[width=7cm]{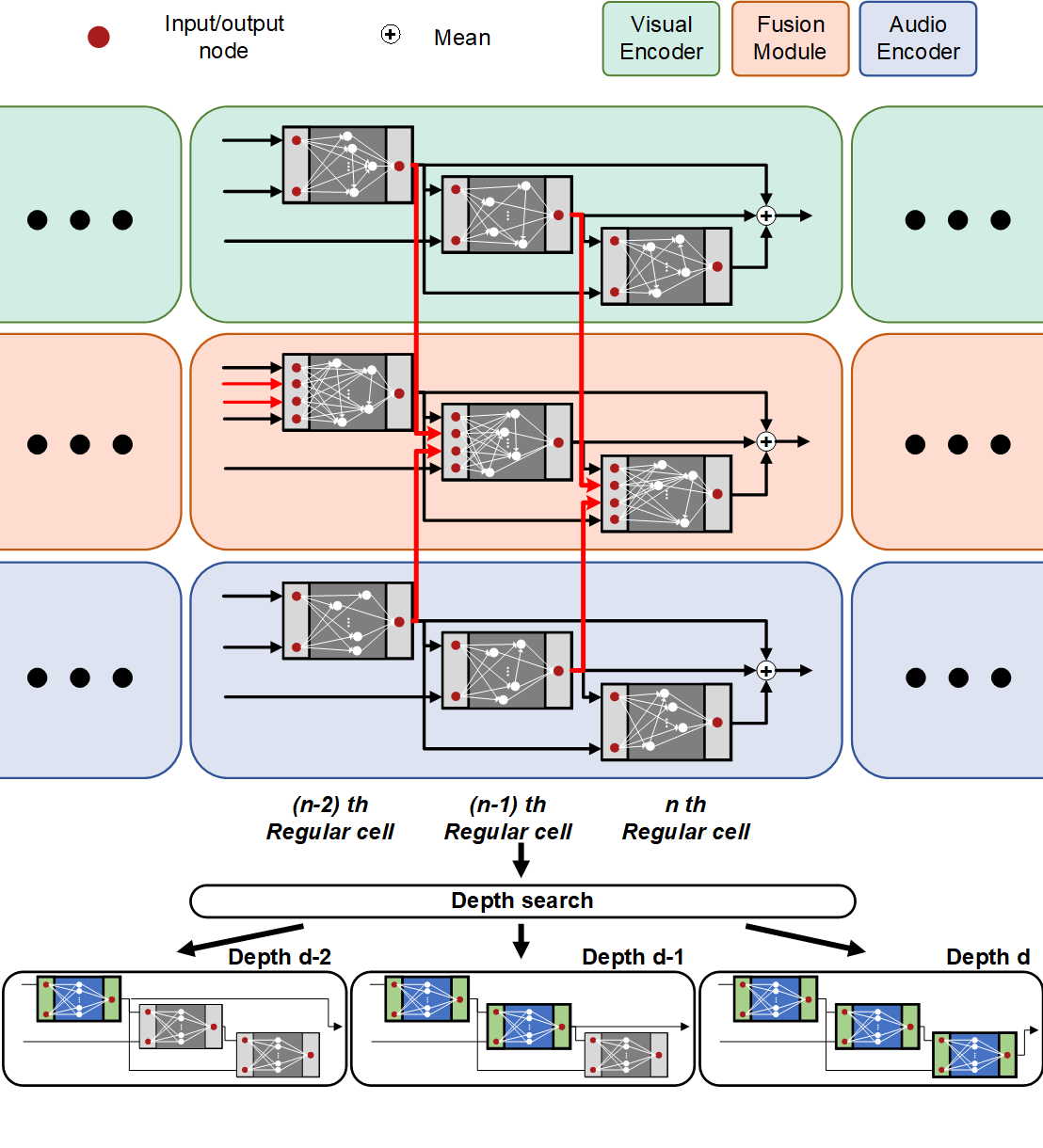}}
	\subfigure[The details of the internal cell connections. Each node in a regular cell is influenced by the information coming from all its previous nodes, and each regular cell takes the outputs from previous two regular cells.]{\label{subfig:details}
		\includegraphics[width=7cm]{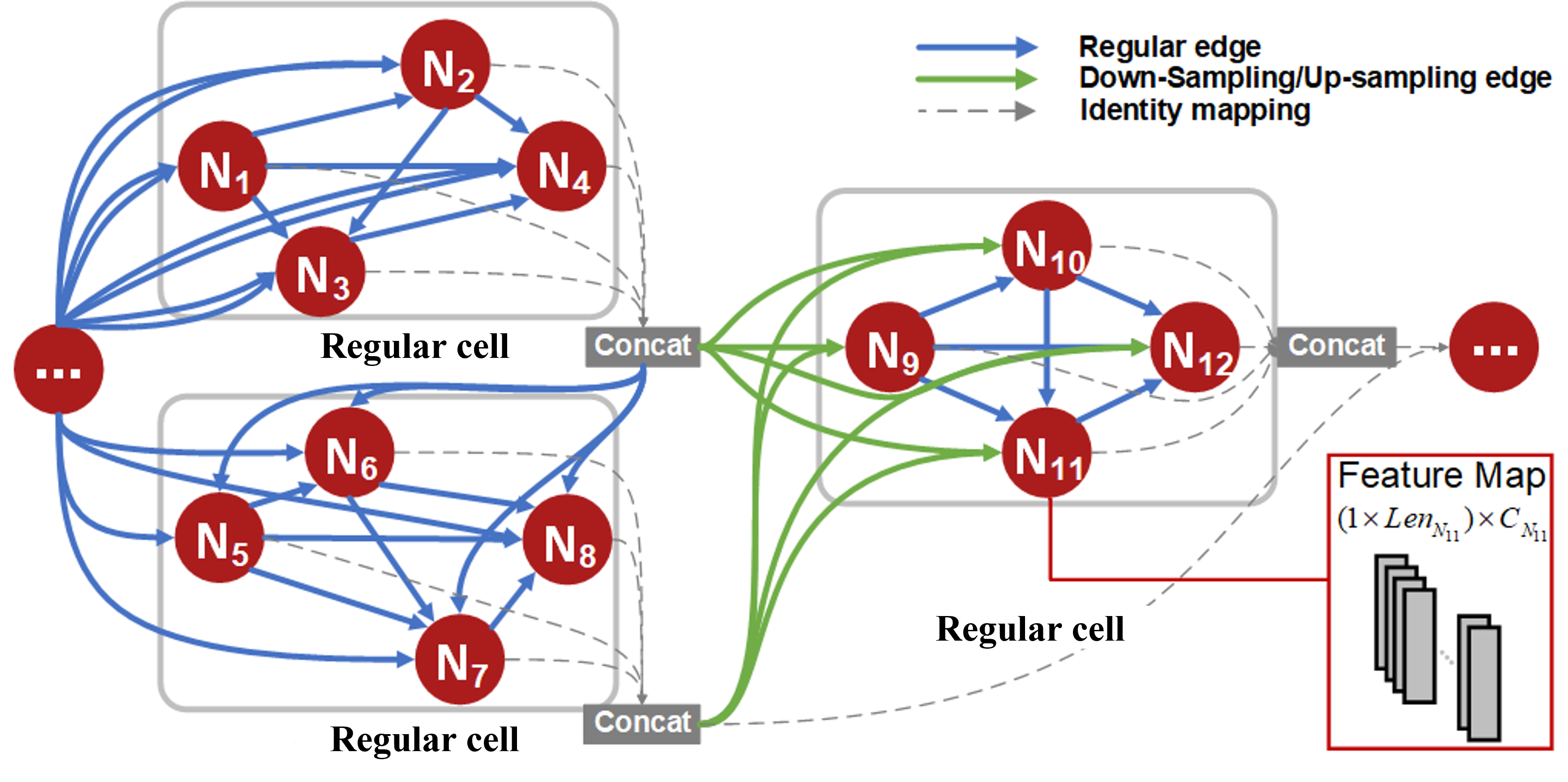}}
	\caption{The illustration of the network details.} 
    \label{fig:network_details}
\end{figure}

We follow previous NAS approaches \cite{liu2018darts,pham2018efficient} to represent each CNN architecture as a directed acyclic graph that contains two basic elements: (i) nodes and (ii) directed edges. In our setting, a node $N_j$ represents a set of feature maps generated from its adjacent parent nodes $N_{i}, N_{i+1}, \cdots N_{j-1}$ ($i<j$). A pair of adjacent nodes ($N_i$ and $N_j$) are connected by a directed edge $DE_{i,j}$ which consists of a set of pre-defined operations $o_{i,j}^k$ (e.g., convolution, pooling, etc.). As a result, during the propagation, the feature maps in the node $N_j$ are jointly produced by all operations on all of its adjacent parent nodes, which can be formulated as
\begin{equation}\label{eq:oper}
N_j = \sum_{i<j} (O_{i,j}(N_i) \times \text{adj}(i,j))
\end{equation}
where $\text{adj}(i,j))$ denotes the binary adjacency relationship (i.e., $0$ denotes non-adjacent and $1$ denotes adjacent) between $N_i$ and $N_j$. Here, each operation $o_{i,j}^k$ in $DE_{i,j}$ is learned to have a unique operation parameter (OP) $\alpha_{i,j}^k$ to represent its importance. As a result, when feeding feature maps of the node $N_i$ to a directed edge $DE_{i,j}$, the output $O_{i,j}(N_i)$ can be represented as:
\begin{equation}\label{eq:oper_k}
O_{i,j}(N_i) = \sum_{k=1}^{K}(\alpha_{i,j}^k \times o_{i,j}^k(N_i))
\end{equation}
In our work, we pre-define $\upsilon = 5$ operations that have LWs, and $\kappa = 5$ operations that do not have LWs, which are listed in Table. \ref{tb:operation}. To simulate the uncertainty and complexity of human cognitive and reaction processes, firstly, the $n_{th}$ ($n>2$) fusion block takes four inputs: the outputs of the $n_{th}$ visual block and $n_{th}$ audio block, the output of the $(n-1)_{th}$ fusion block, and the output of the $(n-2)_{th}$ fusion block. Consequently, the input audio and visual signals can be combined and jointly processed at multiple levels (illustrated in Fig. \ref{subfig:P-CNN}). Secondly, in each regular cell, we set each node to connect to all of its previous nodes to represent all possible information flow, allowing the extracted features (nodes) to be potentially influenced by the information of multiple previous states (nodes) during the CNN learning (illustrated in Fig. \ref{subfig:details}). Thirdly. we set each CNN edge to have a set of unique OPs and LWs rather than setting all cells to share the same set of OPs \cite{liu2018darts,pham2018efficient}.



In addition to the OPs and LWs, depth is also a key factor that impacts a CNN's inference process. Thus, the depths of optimal person-specific CNNs used for representing different subjects' cognitive processes are unlikely to be the same. This paper also proposes to assign each regular block of each person-specific CNN to have a unique depth. For a person-specific CNN, we search for a unique number of cells for each regular block. Specifically, we initially set each regular block to contain $N_{reg}$ linear stacked cells, where each cell produces two outputs: an output which is used as the input to the next cell, and an output that contributes to the entire block's output. Inspired by \cite{wang2020rethinking}, during a person-specific CNN search, we individually mask out each cell of each regular block. Specifically, when masking out the $m_{th}$ cell in a regular block, all its succeeding cells (i.e., $m+1_{st}, m+2_{nd}, \cdots$, $N_{th}$) are also blocked, i.e., the output of the block is decided by the sub-network that is made up of $m$ cells ($1_{st}, 2_{nd}, \cdots$, $m_{th}$). The performance of the CNN is measured after masking out each cell. If masking out the $m_{th}$ cell leads to the largest drop in facial reaction generation performance, the final optimal depth for the corresponding block is $m$.

\begin{table}[t]
\begin{center}
\normalsize
\begin{tabular}{|  c | c| c|}
\hline
Operation Name & Size & Num of weights \\
\hline
Max Pooling & $1 \times 3$ & 0 \\
Average Pooling & $1 \times 3$ & 0 \\
Separable Convolution & $1 \times 3$ & $3 \times C_{in} \times C_{out}$ \\
Separable Convolution & $1 \times 5$ & $5 \times C_{in} \times C_{out}$  \\
Dilated  Convolution & $1 \times 3$ & $3 \times C_{in} \times C_{out}$  \\
Dilated  Convolution & $1 \times 5$ & $5 \times C_{in} \times C_{out}$  \\
Transposed Convolution & $1 \times 3$ & $3 \times C_{in} \times C_{out}$ \\
Up-Sampling (Linear) & N.A. & 0 \\
Up-Sampling (Nearest) & N.A. & 0 \\
Identity Mapping & N.A. & 0  \\
\hline
\end{tabular}
\end{center}
\caption{The operations used in this paper. $C_{in}$ and $C_{out}$ denote the numbers of input and output feature maps, respectively.}
\label{tb:operation}
\end{table}

\subsubsection{Adaptive loss function} 
\label{subsec: adaptive loss}

\noindent This section proposes a novel adaptive loss function to evaluate the difference between the predicted facial reactions and the ground-truth of the target listener's real facial reactions, in order to supervise person-specific CNNs' searching process (i.e., achieving Eqa. \ref{hypo}). Firstly, we notice that facial reactions of similar emotions or intentions can be displayed by different facial spatio-temporal patterns, which is partially caused by the differences in listeners' facial identities and responding times. Such uncertainties may bring bias to the explored CNNs. While differences in facial identities can be partially addressed by projecting faces of different subjects to a mean face (also illustrated in Sec. \ref{subsec: input_target}), we consider that there is always a time delay for a listener to generate a facial reaction, after receiving the speaker's signal. This is because the execution of the corresponding cognitive processes takes some time (as stated in \cite{card1986model}). Importantly, the duration of the time delay may vary not only for different listeners but also for the same listener depending upon other external factors.

In light of this, we introduce an adaptive factor $\tau$ to model this uncertainty. Let us define an audio-visual input $A_{S}(t_1,t_2)$ and $F_{S}(t_1,t_2)$ that represent the speaker's audio-visual non-verbal behaviours produced from time $t_1$ to $t_2$. We propose the following adaptive loss (A-loss) function to measure the similarity between the predicted listener's facial reaction and the ground-truth:
\begin{equation}\label{eq:loss}
\begin{split}
& L_{A-loss}(t_1, t_2, \tau) \\
& = L_{A-loss}(F_{L}^{p}(t_1,t_2), F_{L}^{g}(t_1+\tau,t_2+\tau)) \\
& = \sum_{i = t_1}^{t_2} \sum_{j = 1}^{68} \text{min}(L_{\star}(x_{i,j}^p, x_{i+\tau,j}^g) + L_{\star}(y_{i,j}^p, y_{i+\tau,j}^g)  , \varepsilon)
\end{split}
\end{equation}
where $F_{L}^{p}(t_1,t_2)$ are the predicted listener's facial landmarks corresponding to the input $A_{S}(t_1,t_2)$ and $F_{S}(t_1,t_2)$; $F_{L}^{g}(t_1+\tau,t_2+\tau)$ are the listener's real facial reaction landmarks induced by $A_{S}(t_1,t_2)$ and $F_{S}(t_1,t_2)$, where $\tau$ represents the time delay; $(x_{i,j}^p, y_{i,j}^p)$ denotes the predicted coordinates of the $j$th facial landmark of the $i$th frame and $(x_{i+\tau,j}^g, y_{i+\tau,j}^g)$ is the corresponding ground-truth coordinate. Specifically, the $\varepsilon$ is a constant value employed to avoid extremely large loss values caused by outliers (e.g. incorrectly detected face regions) which can lead to a misguided CNN search. $L_{\star}$ represents the similarity measurement between the prediction and ground-truth. In this paper, $L_{\star}$ is defined as the Mean Square Error (MSE).


To achieve the proposed adaptive loss (i.e., computing a $\tau$ at each time), in practice we use a sliding time-window to compare the prediction of listener's facial reactions with a set of ground-truth candidates (the duration of the ground-truth candidates is longer than the time-window, as illustrated in the last section of Fig. \ref{subfig:P-CNN}). Specifically, we set $R$ ground-truth candidates, i.e., $F_{L}^{g}(t_1+r,t_2+r), r = 1, 2, \cdots R$, and only choose $r = \tau$ that allows the loss $L_{A-loss}(t_1, t_2, \tau)$ to have the lowest value:
\begin{equation}\label{eq:delay}
\tau =  \text{argmin} L_{A-loss}(t_1, t_2, \tau)
\end{equation}
As a result, the delay period can be automatically adapted for each listener at each training iteration.

\begin{figure*}
\centering
\includegraphics[width=16cm]{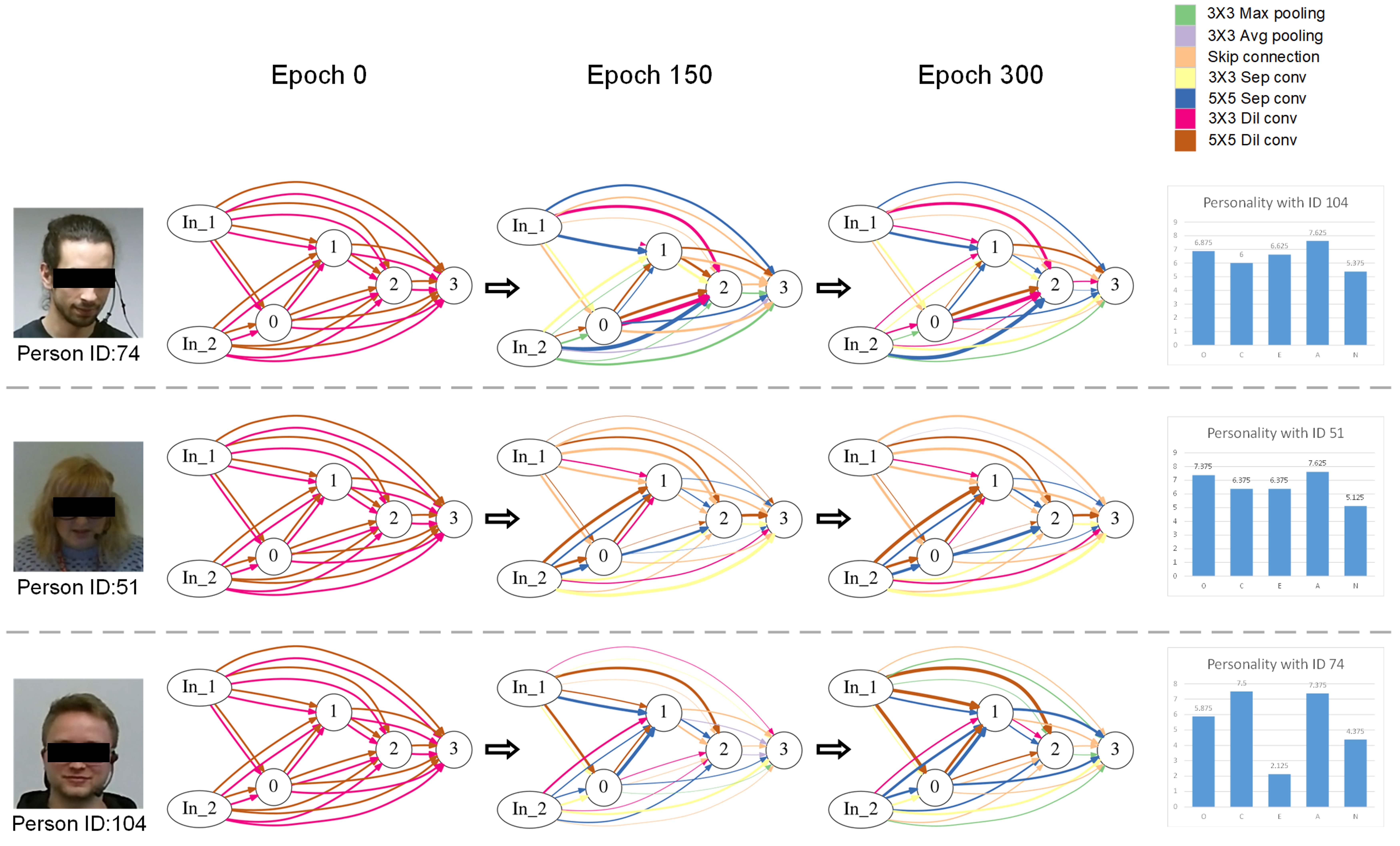}
\caption{Visualization of the person-specific CNN search on the NoXI dataset, where initial CNN architectures for all subjects are the same (Epoch 0). After the CNNs' search, we can see that the explored CNN for each subject is unique and person-specific (Epoch 300).}
\label{fig:visualize}
\end{figure*}

\subsubsection{Person-specific CNN architecture optimization} 
\label{subsec: optimization}

\noindent In this paper, we independently search for an optimal multi-modal CNN architecture (described in Sec. \ref{subsec: multimodal}) for each listener. To minimize the proposed adaptive loss (Eqa. \ref{eq:loss}), we conduct a single-level optimization using the continuous relaxation algorithm \cite{liu2018darts}.

As illustrated in Algorithm 1, the proposed strategy adjusts both OPs and LWs of an entire CNN at the same time during the optimization. In comparison to the widely-used bi-level optimization strategy which separately optimizes OPs in the validation set and LWs in the training set, i.e., freezing one of them while optimizing the other, the proposed single-level optimization strategy allows the OPs and LWs to be simultaneously optimized. This aims to replicate how the human cognition operates with all cognitive processes jointly activated during reaction generation \-- there is no evidence suggesting that some parts of the human model processor are frozen during the reaction generation. In addition, this strategy allows the OPs and LWs to be optimized using the full audio-visual clip instead of a sub-segment of it, i.e., the explored CNN is a video-level representation without ignoring any frames. Representative examples of person-specific CNNs' optimization processes are visualized in Fig. \ref{fig:visualize}.

\begin{algorithm}[t]
\label{alg:single}
  \caption{Single-level optimization}  
  \begin{algorithmic}[1]
    \Require 
      A multi-modal CNN architecture parametrized by OPs and LWs of an audio encoder, a visual encoder, a fusion module and a decoder, denoted as $\alpha_{V,A,F,D}^{t=0}$ and $\omega_{V,A,F,D}^{t=0}$.
      
    \Ensure  
      A optimal person-specific multi-modal CNN that can reproduce the target subject's facial reactions, which is parametrized by OPs and LWs, denoted as $\alpha_{V,A,F,D}^{\text{Optimal}}$ and $\omega_{V,A,F,D}^{\text{Optimal}}$.
      
      \Repeat
      
        \State Update LWs $\omega_{V,A,F,D}^t$ by descending $L_{A-loss}(\omega_{V,A,F,D}^{t-1} - \eta_{\omega} \nabla L_{A-loss}(\omega_{V,A,F,D}^{t-1}(\alpha_{V,A,F,D}^{t-1}), \alpha_{V,A,F,D}^{t-1})$
        
        \State Update OPs $\alpha_{V,A,F,D}^t$ by descending $L_{A-loss}(\alpha_{V,A,F,D}^{t-1} - \eta_{\alpha} \nabla L_{A-loss}(\alpha_{V,A,F,D}^{t-1}(\omega_{V,A,F,D}^{t-1}), \omega_{V,A,F,D}^{t-1})$
        
      \Until{Convergence}
      
      \State $\alpha_{V,A,F,D}^{\text{Optimal}} = \alpha_{V,A,F,D}^{\text{Convergence}}$ and $\omega_{V,A,F,D}^{\text{Optimal}} = \omega_{V,A,F,D}^{\text{Convergence}}$
      
  \end{algorithmic}  
\end{algorithm}

\subsection{Graph representation of the person-specific CNN architecture}
\label{subsec: graph representation}

\noindent One of our main hypothesis in this paper is that a CNN that has been trained to reproduce the target subject's facial reactions, can be used to represent their cognitive processes. Therefore, the CNN's architectural parameters (OPs and LWs) and depth, which decide its cognitive process, should be informative for modeling the target subject's true personality. To allow for the person-specific CNN to be modelled by standard ML regression techniques, we encode each CNN's architectural parameters and depths into a graph representation $G(V,E)$ that is made up of a set of vertices and edges, as CNNs have graphical topology.

\subsubsection{Vertex feature encoding}
\label{subsec: vertex feature}

\noindent We represent a directed edge $DE_{i,j}$ in the explored CNN as a vertex $V_{i,j}$ in the corresponding graph representation $G(V,E)$. This process is illustrated in Fig. \ref{subfig:Graph} and is depicted in purple. In particular, the $DE_{i,j}$ is made up of a set of learned OPs $\alpha_{i,j}^k$ and LWs $\omega_{i,j}^k$, $k = 1, 2, \cdots K$, where $\alpha_{i,j}^k$ and $\omega_{i,j}^k$ represent the OP and LWs of the $k_{th}$ operation in $DE_{i,j}$.

We first notice that the number of valid operations and their LWs partially depend on the type of the target directed edge. For example, the up-sampling and down-sampling operations are not used in directed edges of regular blocks while the transposed convolution is not used for directed edges of down-sampling blocks. Besides, they also depend on the number of input and output feature maps, which can be different for different directed edges in a CNN (ranging from hundreds to tens of thousands in our study). To allow each vertex feature to have the same dimensionality, we first set all directed edges to have the OPs of all pre-defined operations ($10$ in this paper), where the OPs of invalid operations are set to zero. Then, we propose two solutions to encode the arbitrary number of LWs of a directed edge $DE_{i,j}$ into a LWs representation $S\omega_{i,j}$ with a fixed dimension: firstly. we follow the idea of \cite{li2016pruning} to select a subset of weights $S\omega_{i,j}^k$ from each operation $o^k$ to represent its most important kernels, i.e., we only choose kernels with the top-5 highest L1 values (sum of absolute weights), and use their weights as LWs representation; secondly. for each set of kernels corresponding to each operation $o^k$, we construct a histogram to represent the distribution of weights at each position in the kernels. For example, to summarize the weights of a set of $1 \times 3$ kernels, $3$ histograms are constructed corresponding to all positions in the kernels, regardless of the number of kernels. These histograms are concatenated and used as a representation for LWs.


To combine the produced OPs and LWs representation described above, we notice that each OP $\alpha_{i,j}^k$ reflects the importance of the corresponding operation which contains the LWs representation $S\omega_{i,j}^k$. We categorize all operations into two parts: $\upsilon$ operations that have LWs (e.g., convolution) and $\kappa$ operations that do not have LWs (e.g., pooling). For operations that have LWs, their original OPs $\alpha_{i,j}(w)$ are projected to a OP-LW weighting vector represented as:
\begin{equation}
OP^{LW}_{i,j} = \text{VEN}(\alpha_{i,j}(w), \omega_{\text{VEN}})
\end{equation}
Here $OP^{LW}_{i,j}$ has the same number of dimensions as $S\omega_{i,j}$, where VEN is a vertices features encoding network and $\omega_{\text{VEN}}$ denotes the learnable weights of the VEN. The representation for operations that have LWs is obtained by computing the dot product between the weighting vector $OP^{LW}_{i,j}$ and the LWs representation $S\omega_{i,j}^k$ as:
\begin{equation}
\label{eq:op-lw}
V_{i,j}(\omega) = \langle OP^{LW}_{i,j}, S\omega_{i,j} \rangle
\end{equation}
In this paper, all vertices share a single VEN for their features' encoding, which specifically learns the importance of each LW based on not only its corresponding OP but also the relationship between the LW and the target (personality traits in this paper). Finally, we concatenate the obtained $V_{i,j}(\omega)$ with other $\kappa$ OPs of operations that do not have LWs, as the final vertex feature:
\begin{equation}
\label{eq:vector}
V_{i,j} = [\alpha_{i,j}^{kn_1},\alpha_{i,j}^{kn_2}, \cdots, \alpha_{i,j}^{kn_\kappa}, V_{i,j}(\omega)]
\end{equation}
where $kn$ denotes operations that do not have LWs. This process is described in Algorithm 2.

\begin{algorithm}[t]
\label{alg:alg-vertex}
  \caption{Vertex feature encoding}  
  \begin{algorithmic}[1]
    \Require 
      A directed edge $DE_{i,j}$ in the searched CNN, which contains:
      \textbf{(1)}. $o_{i,j}^{kw}$ ($kw = kw_1,kw_2,\cdots,kw_{\upsilon}$): the $kw_{th}$ operation, which has layer weights; 
      \textbf{(2)}. $o_{i,j}^{kn}$ ($kn = kn_1,kn_2,\cdots,kn_{\kappa}$): the $kn_{th}$ operation, which does not have layer weights; 
      \textbf{(3)}. $\alpha_{i,j}^{kw}$: The OP of $kw_{th}$ operation;   
      \textbf{(4)}. $\alpha_{i,j}^{kn}$: The OP of $kn_{th}$ operation;   
      \textbf{(5)}. $\omega_{i,j}^{kw}$: The layer weights of $kw_{th}$ operation.
      
    \Ensure  
      A vertex feature $V_{i,j}$ that has a fixed dimension.

      \Repeat 
      
      \State Producing a LW representation $S\omega_{i,j}^{kw}$ for $o_{i,j}^{kw}$.
       
      \Until{($kw = kw_{\upsilon}$)}

      \State Concatenating OPs that have LWs as: $\alpha_{i,j}(w) \leftarrow [\alpha_{i,j}^{kw_1}, \alpha_{i,j}^{kw_2}, \cdots, \alpha_{i,j}^{kw_{\upsilon}}]$
      
      \State Feeding $\alpha_{i,j}(w)$ to VEN to generate a LWs weighting vector: $V_{i,j}(\alpha_{i,j}(w)) \leftarrow  \text{VEN}(\alpha_{i,j}(w), \omega_{\text{VEN}})$
      
      \State Concatenating all LWs' representations as: $S\omega_{i,j} \leftarrow [S\omega_{i,j}^{kw_1}, S\omega_{i,j}^{kw_2}, \cdots, S\omega_{i,j}^{kw_{\upsilon}}]$
      
      \State Weighting  $S\omega_{i,j}$ to produce a representation for operations that have LWs: $V_{i,j}(w) \leftarrow \langle V_{i,j}(\alpha_{i,j}(w)), S\omega_{i,j} \rangle$
      
      \State Concatenating all OPs that do not have LWs as a vector $V_{i,j}(n)  \leftarrow [\alpha_{i,j}^{kn_1},\alpha_{i,j}^{kn_2}, \cdots, \alpha_{i,j}^{\kappa}]$
      
      \State Concatenating $V_{i,j}(w)$ and $V_{i,j}(n)$ to generate the vertex feature: $V_{i,j} \leftarrow  [\alpha_{i,j}^{kn_1},\alpha_{i,j}^{kn_2}, \cdots, \alpha_{i,j}^{kn_\kappa}, V_{i,j}(w)]$
      
  \end{algorithmic}  
\end{algorithm}

\subsubsection{Edge feature encoding}
\label{subsec:edge_feat}

\begin{figure*}
\centering
\includegraphics[width=16cm]{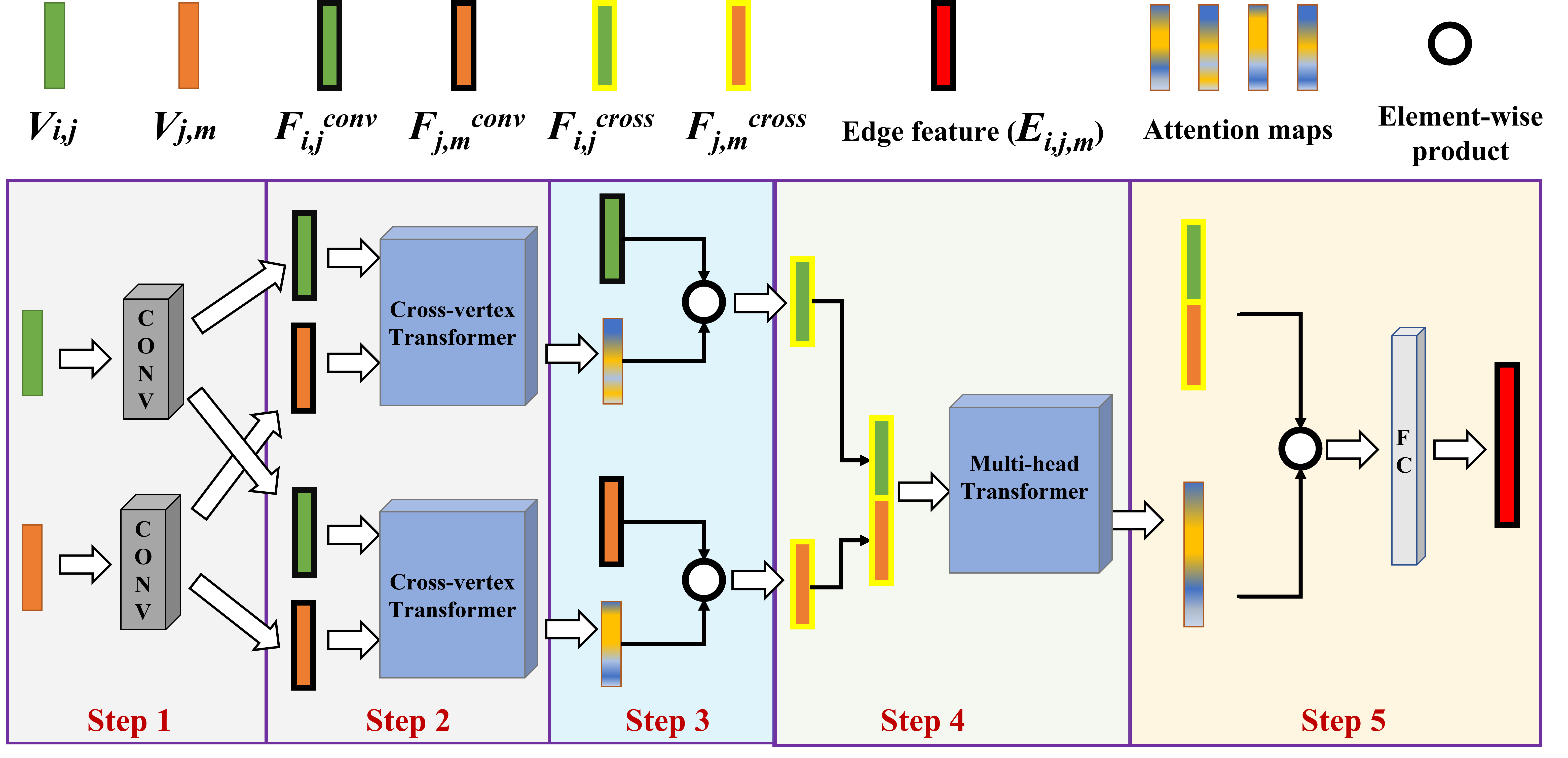}
\caption{Illustration of the ERN used for edge feature encoding. Step 1: Learning a pair of 1D representations $F_{i,j}^{conv}$ and $F_{j,m}^{conv}$ from original vertices features; Step 2: Learning cross-vertex attention maps $A_{i,j}^{cross}$ and $A_{j,m}^{cross}$; Step 3: Generating weighted feature representation $F_{i,j}^{cross}$ and $F_{j,m}^{cross}$; Step 4: Concatenating $F_{i,j}^{cross}$ and $F_{j,m}^{cross}$, and learning self-attention maps; and Step 5: Generating the edge feature $E(i,j,m)$.}
\label{fig:edge_feat}
\end{figure*}

\noindent For a graph representation $G(V,E)$, we define that a pair of vertices $V_{i,j}$ and $V_{j,m}$ are adjacent if their corresponding directed edges $DE_{i,j}$ and $DE_{j,m}$ are connected to the same node $N_j$ in the CNN (illustrated in Fig.\ref{subfig:Graph} in yellow). Instead of defining that all vertices are connected, this setting not only prevents the produced graph representation from having an overly large number of edges but also retains the original adjacency among directed edges in the explored CNN. While most GNN-based approaches used a single binary adjacency value ($0$ or $1$) to define the relationship between a pair of vertices, valuable relationship clues may be ignored. Thus, we propose to use multi-dimensional edge features to represent the relationship between each pair of vertices in our graph representations.

Inspired by the fact that all directed edge's OPs and LWs are jointly optimized during the search process of persons-specific CNN architectures, we define $E(i,j,m)$ as the relationship between a pair of adjacent directed edges $DE_{i,j}$ and $DE_{j,m}$ during the optimization process, i.e., how the changes in a directed edge's OPs and LWs influence the OPs and LWs of an adjacent directed edge. Thus, $E(i,j,m)$ can be used as the edge feature for vertices $V_{i,j}$ and $V_{j,m}$ in the graph representation. In particular, we propose to learn $E(i,j,m)$ directly from $V_{i,j}$ and $V_{j,m}$:
\begin{equation}\label{eq:edge_feat}
E(i,j,m) = \text{ERN}_{i,j,m}(V_{i,j}, V_{j,m}, \omega_{\text{ERN}_{i,j,m}}))
\end{equation} 
where $\omega_{\text{ERN}_{i,j,m}}$ denotes the learnable weights of a edge relationship network $\text{ERN}_{i,j,m}$ that is specifically trained to define the relationship between $V_{i,j}$ and $V_{j,m}$.

In this paper, each ERN is a transformer that takes a pair of vertices' feature $V_{i,j}$ and $V_{j,m}$ as the input and output the edge feature $E(i,j,m)$. Each ERN consists of three main blocks: (i) a two-stream convolution block that generate a pair of 1D representations $F_{i,j}^{conv}$ and $F_{j,m}^{conv}$ that are learned from $V_{i,j}$ and $V_{j,m}$, respectively; (ii) a two-head cross-vertex attention module that generates a pair of attention features $F_{i,j}^{cross}$ and $F_{j,m}^{cross}$ from $F_{i,j}^{conv}$ and $F_{j,m}^{conv}$. In particular, this module learns the relationships from two vertices' perspective.  Firstly, $F_{j,m}^{conv}$ is used as the query and $F_{i,j}^{conv}$ is treated as the key to learn an attention map $A_{i,j}^{cross}$ that emphasizes a part of $V_{i,j}$'s information correlated with $V_{j,m}$. Secondly, $F_{i,j}^{conv}$ is used as the query and $F_{j,m}^{conv}$ is treated as the key to learn an attention map $A_{j,m}^{cross}$ that emphasizes a part of $V_{j,m}$'s information correlated with $V_{i,j}$. Then, these learned attention maps are employed to weight the $F_{i,j}^{conv}$ and $F_{j,m}^{conv}$ to generate weighted feature representations $F_{i,j}^{cross}$ and $F_{j,m}^{cross}$ where mutually-related features are highlighted; (iii) the third block is a multi-head transformer module that contains a self-attention layer and a fully connected layer, which further extracts the most important relationship features $E(i,j,m)$ that have been emphasized by both cross-vertex features (e.g., $F_{i,j}^{cross}$ and $F_{j,m}^{cross}$), whose input is a vector that concatenates both cross-vertex attention features. The proposed ERN network is motivated by the transformer described in \cite{tsai2019multimodal}, which has shown excellent performance on multi-modal tasks. The full pipeline of this end-to-end edge feature encoding process is depicted in Fig. \ref{fig:edge_feat}. It should be noted that all ERNs are jointly trained with the personality recognition model in an end-to-end manner. This way, all ERNs are learned to generate task-specific edge features (personality-related features) for the corresponding vertices pairs.




      
     
      
     
     
     
     
      

\subsubsection{Graph topology alignment}    
\label{subsec: vertex feature}

\noindent As described in Sec. \ref{subsec: multimodal}, the explored person-specific CNNs may have different depths, i.e., different numbers of cells in regular blocks. While we can directly feed the heterogeneous graph representations of person-specific CNNs to some advanced GCNs that can process heterogeneous graphs, this paper also proposes two solutions to parameterize variable-depths CNNs into graph representations with the same numbers of vertices and edges:

\begin{itemize}
    
    \item \textbf{Block distillation:} For a regular block with an arbitrary depth, we distill its cognitive processes into a regular block that has the fixed depth. In other words, we distill all regular blocks with different depths into a set of regular blocks that have the same depth. As a result, all person-specific CNNs as well as their graph representations would have the same topology.

    \item \textbf{Block Maximization:} We assume that all regular blocks of all person-specific CNNs have the maximum number of cells. Then, for those regular blocks that were not optimized to have the maximum number of cells during the CNN search, we set the redundant cells as the identity cells. In other words, in each redundant cell, only one identity mapping operation's OP is set to $1$ while the rest of OPs and LWs are all set to 0. This way, graph representations of all person-specific CNNs would have the same topology, regardless of their differences in depth.

\end{itemize}

\subsection{Personality recognition model}

\noindent In this paper, the target subjects' personality traits are recognised from the graph representations of their person-specific CNNs. In this sense, we formulate the personality recognition as a multi-task graph regression problem  (recognizing $5$ traits). Particularly, we employ the state-of-the-art residual gated graph convolution neural network (residual GatedGCN) \cite{bresson2017residual} as the personality recognition model to process the produced graph representations, which can jointly process heterogeneous graphs. We empirically employ a network that consists of six GatedGCN layers. Then, two fully connected (FC) layers are attached to the last GatedGCN layer to concatenate all produced vertices features, where a ReLU activation and a dropout ($0.3$) are followed by each FC layer. The size of the output layer is set to $5$ to jointly recognise the \textit{five} personality traits of \textbf{Extroversion} (\textbf{Ext}), \textbf{Agreeableness} (\textbf{Agr}), \textbf{Openness} (\textbf{Ope}), \textbf{Conscientiousness} (\textbf{Con}), and \textbf{Neuroticism} (\textbf{Neu}).

\section{Experiments}

\noindent This paper evaluates the proposed approach on a human-human dyadic interaction dataset and a human-machine dyadic interaction dataset, which are described in Sec. \ref{subsec:database}. We then present the implementation details in Sec. \ref{subsec:imple} followed by evaluation metrics in Sec. \ref{subsec:evaluate}. We compare the personality recognition performance achieved by the proposed approach to other existing personality recognition approaches in Sec.\ref{subsec:sota}, showing the advantages of the proposed novel strategy which recognises personality traits from the simulated human cognitive processes. Moreover, we conduct a set of additional experiments to systematically evaluate the capability and sensitivity of our approach for different demographic groups in Sec. \ref{subsec:interactive} as well as the influence of different model settings of the person-specific CNN search and graph representations on personality recognition in Sec. \ref{subsec:ablation}.

\subsection{Datasets}
\label{subsec:database}

\noindent In this paper, we evaluate our approach in both human-human and human-machine dyadic interaction scenarios. While many existing datasets \cite{mckeown2012semaine,ponce2016chalearn,biel2010voices,sanchez2011nonverbal} are built for personality perception prediction study, some publicly available datasets \cite{jaiswal2019automatic,correa2018amigos,palmero2021context,cafaro2017noxi,celiktutan2017multimodal} also can be used for audio-visual true personality recognition studies. However, most of these available datasets are not suitable for our study as our models assume dyadic interaction.


\textbf{Human-human interaction:} The NoXi dataset \cite{cafaro2017noxi} is a multi-lingual human-human dyadic interaction dataset that was designed to generate spontaneous interactions with emphasis on adaptive behaviours in unexpected situations. The dataset consists of $84$ sessions in which one participant acts as an Expert and the other acts as a Novice interacting on a chosen topic of expertise via video conferences. The participants were allowed to continue the conversation until it reached a natural end. During the interaction, participants can interrupt each other for either changing the topic or inducing a mild debate whenever possible. The NoXi dataset contains $84$ pairs of audio-visual clips ($168$ clips in total from $89$ participants) with participants' ages ranging from 21 to 50 years. The average and standard deviation of the clips' duration are $18$ mins $6$ seconds and $6$ mins $28$ seconds, respectively. All participants provided the self-assessments of their Big-Five Personality Traits using the Saucier’s Mini-Markers \cite{saucier1994mini}.

\begin{figure}
\centering
\includegraphics[width=8cm]{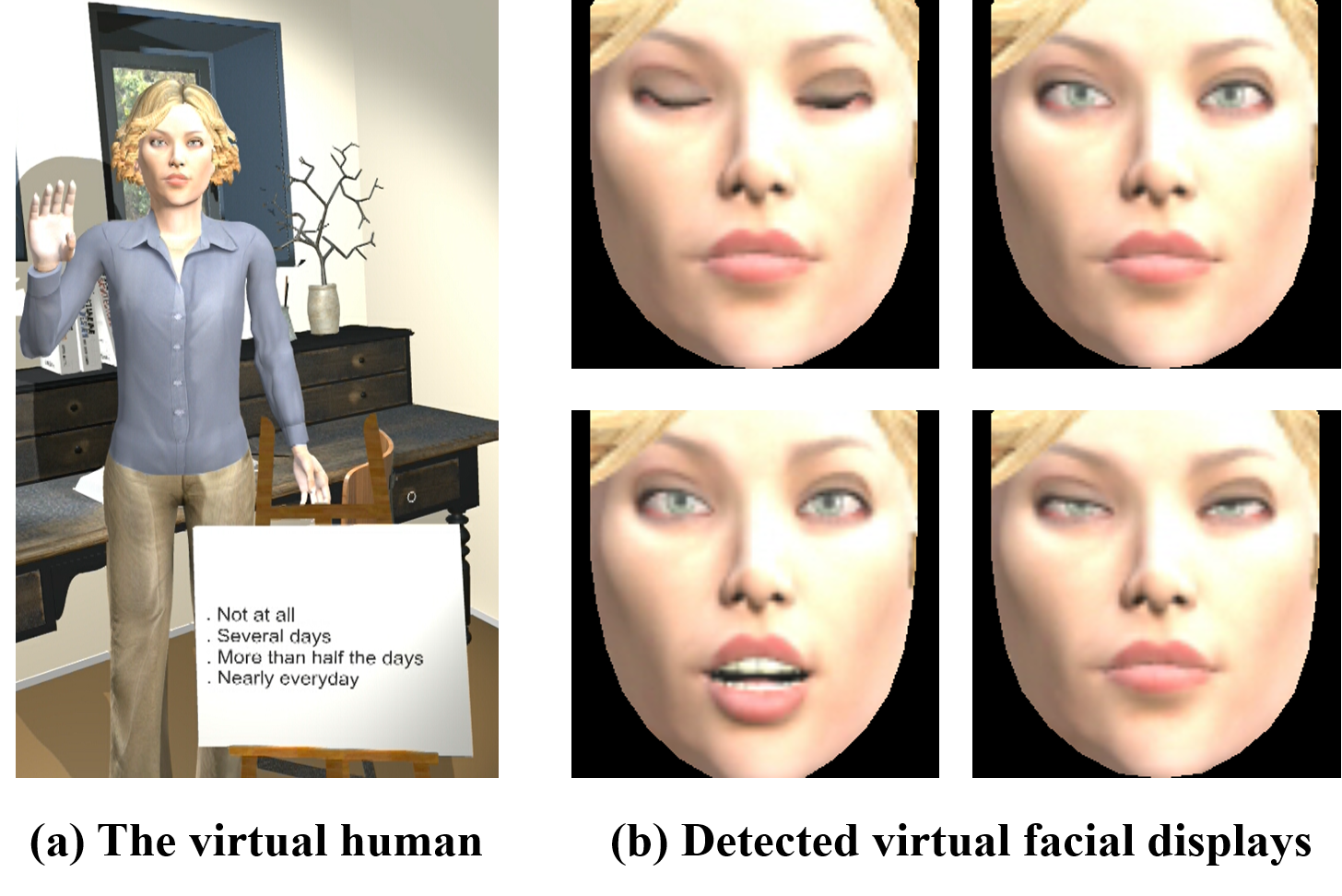}
\caption{Examples of a virtual human display and automatically detected (aligned) faces in VHQ dataset.}
\label{fig:virtual_human}
\end{figure}

\textbf{Human-machine interaction:}  We conducted the human-machine experiments on the Virtual Human Questionnaire (VHQ) database \cite{jaiswal2019virtual,jaiswal2019automatic}. The VHQ database consists of $165$ videos collected from $55$ participants, where each participant completed $3$ questionnaire interview sessions. During each session, participants were asked to answer a set of questions verbally based on one of three questionnaires: BFI-10 \cite{rammstedt2007measuring}, PHQ-9 \cite{kroenke2002phq} or GAD-7 \cite{spitzer2006brief}. In this database, $55$ videos (corresponding to $55$ subjects) were recorded under the human-machine dyadic interaction mode. More specifically, a virtual human agent interviewer (Fig. \ref{fig:virtual_human}) was projected directly in front of the participant and ask questions, which was implemented using the ARIA-VALUSPA Platform \cite{avp2019}. The self-reported labels of the Big-Five personality traits were obtained by asking participants to fill the BFI-44 questionnaire online.

\subsection{Implementation details}
\label{subsec:imple}


\textbf{Neural architecture search:} For subjects in the NoXI dataset, their multi-modal person-specific CNNs have $6$ blocks (a pre-defined convolution block, $3$ down-sampling blocks and $2$ regular blocks) for each encoder and $5$ blocks ($2$ regular blocks and $3$ up-sampling blocks) for each decoder. The employed LSTMs have $3$ hidden layers. During the neural architecture search, the input speaker's audio-visual signal lasts for $80$ frames and the listener's candidate ground-truth consists of $105$ frames and the delay factor $r$ ranges from $0$ to $25$ frames, i.e., selecting $80$ consecutive frames as the final reaction. In our experiment, the batch size was set to $60$ audio-visual clips, while $2$ Adam optimizers were independently used to jointly adjust OPs $\alpha$ and LWs $\omega$, with the learning rate of $0.05$ and $0.001$, respectively.

Since the audio data in the VHQ dataset is very noisy, we only search for a single-modal person-specific CNN for each subject, which takes the virtual human's facial landmarks as the input and aims to reproduce the target subject's facial reactions. We noticed that the virtual human only spoke a set of pre-defined sentences during the interaction. Thus, we also categorized all sentences into $4$ classes: depression-related questions, anxiety-related questions, personality-related questions, and other sentences (e.g., virtual human asks the subject to repeat the answer.), and then encoded each of them as a four-digit one-hot vector (e.g., $1000$, $0100$, $0010$, and $0001$). In this paper, the multi-modal models used for the VHQ dataset are implemented by concatenating the deep-learned virtual human's face feature with the proposed sentence categorical feature at the last FC layer of the encoder.

\textbf{VEN and ERNs:} The VEN used in our experiments is a Multi-Layer Perceptron (MLP) with two hidden layers (each made up of $60$ neurons), where the size of its input layer equals the number of operations that have LWs ($\upsilon$ in this paper) and the size of the output layer equals the dimension of the corresponding LWs representation. All ERNs have the same architecture (introduced in Sec. \ref{subsec:edge_feat}). In this paper, we set all cells of the same type in a module to share the same set of ERNs, where all ERNs have the same architecture as explained in Sec. \ref{subsec:edge_feat}. For example, all regular cells in the visual encoder share the same set of ERNs, which are different from the ERNs for regular cells in the audio encoder or the fusion module. For each cell, the number of ERNs equals the number of its adjacent vertices pairs (i.e., directed edges) and each ERN specifically learns a feature for its target edge in the cell. This setting avoids training too many ERNs for a person-specific CNN, i.e., it avoids to optimizing an extremely large number of weights. The weights of all ERNs are initialized with the Xavier strategy \cite{glorot2010understanding}.

\textbf{Graph vertex and edge feature setting:} We evaluated the proposed graphical representations under five different vertex feature settings: OPs, LWs, and three OP-LW features (OP-LW(C), OP-LW(W) and OP-LW(VEN)). The OP vertex feature contains $\upsilon + \kappa$ OPs; the LWs vertex feature contains all produced LWs $S\omega_{i,j}$; the OP-LW(C) vertex feature is produced by the simple concatenation of $\upsilon + \kappa$ OPs and the LWs representation; the OP-LW(W) feature is obtained by concatenating OPs that do not have LWs, and a weighted OP-LW vector that is produced by multiplying each component of the LWs representation with its corresponding OP \cite{shao2021personality}; and the OP-LW(VEN) vertex feature is generated by the proposed end-to-end learning approach (explained in Sec. \ref{subsec: vertex feature}). In our experiments, we set the edge features to have the same dimension as the vertex features.

\textbf{GNN-based personality recognition model:} For all experiments, the size of the first GatedGCN layer is set to be the same as the dimensionality of the input vertex feature $d_{in}$. The sizes of the rest of the five GatedGCN layers were set to be $\upsilon + \kappa$, $10$, $10$, $5$ and $5$ for OP features. For LW feature and OP-LW combination features, these were set to $\lfloor d_{in}/2\rfloor$, $\lfloor d_{in}/4\rfloor$, $\lfloor d_{in}/8\rfloor$, $\lfloor d_{in}/16\rfloor$ and $\lfloor d_{in}/32\rfloor$, where $\lfloor\rfloor$ denotes the floor function. The weights of all our GCN models are randomly initialized. In this paper, all experiments were conducted on the PyTorch platform using Nvidia V100 GPUs.

\textbf{Personality model training details:} In this paper, we conduct a 7-fold subject-independent cross-validation on the NoXI dataset. For each fold, $154$ videos were used for training and hyperparameter optimisation and $14$ videos were used for testing (each subject appeared in either training or test set, not both). Due to the limited number of data, we conduct a leave-one-subject-out cross-validation on the VHQ dataset. For each fold, $54$ videos were used for training and hyperparameter optimisation, and the remaining video was used for testing. For both NoXI and VHQ datasets, we report the accuracy on the test sets averaged over all folds.

\textbf{The reproduction of existing approaches:} To compare the proposed approach with other video-based automatic personality analysis solutions, we reproduced several approaches and also evaluated them on the NoXI and VHQ datasets. In particular, we first evaluated all reproduced approaches on the ChaLearn dataset \cite{ponce2016chalearn}, as all of the reproduced approaches have initially reported their results on this dataset. Then, we chose four approaches (e.g., DCC \cite{guccluturk2016deep}, NJU-LAMDA \cite{wei2018deep}, CR-Net \cite{li2020cr}, and PALs \cite{song2021self}) that we have managed to reproduce similar results on the ChaLearn dataset as the baselines on the NoXI and VHQ datasets. In addition, we also employed the spectral representation \cite{song2018human,song2020spectral} as a comparison as it also provides video-level representation and achieved promising results on a video-level human status analysis task (i.e., depression severity estimation). The detailed settings of the reproduced approaches on the NoXI and VHQ datasets are listed below.
\begin{itemize}

    \item \textbf{DCC:} We employed the two-stream ResNet-17 proposed in \cite{guccluturk2016deep} to jointly predict personality traits using each audio-visual frame. The final video-level prediction of each subject was generated by averaging all frame-level predictions. In addition, we followed the same pre-processing strategy in \cite{guccluturk2016deep}, e.g.,  resampled the audio data to $16000$ Hz and the video data with the resolution of $456 \times 256$, etc. During the network training, we also used Adam as the optimizer and the mean absolute error (MSE) as the loss function.

    \item \textbf{NJU-LAMDA:} We employed the DAN$^{+}$ model proposed in \cite{wei2018deep} to obtain frame-level visual predictions for each downsampled video (6 fps), where the Tukey’s biweight loss function \cite{black1996unification} was used to train the visual network. For audio data,  we followed the same setting introduced in \cite{wei2018deep} to extract MFCC features, which are then used to train a linear regressor, where SGD was employed as the optimizer.

    \item \textbf{CR-Net:} We only employed the audio and two-visual streams (e.g., entire images and detected face images) of the CR-Net \cite{li2020cr} in our experiments. Following the same settings as the CR-Net, we downsampled each video into $32$ frames, and using the Bell loss to train the network. Specifically, we replaced the extra tree regressor with the stand MLP to conduct the end-to-end training.

    \item \textbf{PALs:} We followed the same setting (the same network architecture and the rank loss function) described in \cite{song2021self} to learn a person-specific representation ($1984$-D including $992$ weights and $992$ bias) for each listener, which comes from five person-specific adaptation layers. Then, we follow the same settings (e.g., feature selection, ANN models, MSE loss functions, Adam optimizer, etc.) in \cite{song2021self} to train an ANN model to jointly predict five personality traits.

    \item \textbf{Spectral representation:} We set frequency as $256$ for frequency alignment and chose the $80$ lowest frequencies to construct a spectral vector for each listener's video. Then, we follow the same settings (e.g., feature selection, ANN models, MSE loss functions, Adam optimizer, etc.) in \cite{song2021self} to jointly predict five personality traits.

\end{itemize}
In our experiments, it should be noted that while we kept the pre-processing, network architecture, loss function and optimizer to be almost the same as the original papers, the hyper-parameters such as batch size, learning rate, weight decay, etc, were specifically optimized for each method on each dataset (NoXI and VHQ), respectively. In this paper, all experiments were conducted using the PyTorch library\footnote{\url{https://pytorch.org/}}, where DGL library\footnote{\url{https://github.com/dmlc/dgl}} is utilized for GNNs-related operations, i.e., building the personality recognition model for the produced isomorphic graph representations and heterogeneous graph representations.

\subsection{Evaluation metrics}
\label{subsec:evaluate}

\noindent Two common metrics are used to evaluate the personality recognition performance: the Pearson Correlation Coefficient ($\mbox{PCC}$) defined in Eq.~(\ref{eq:cc}), and the \textbf{mean accuracy} measurement ($\mbox{ACC}$, Eq.~(\ref{eq:acc})) which has been adopted in relevant challenge events (e.g., the ChaLearn challenge \cite{ponce2016chalearn}.
\begin{equation}
\mbox{PCC} = \frac{\mathrm{Cov} \left(f,y\right)}{\sigma_f\sigma_y}
\label{eq:cc}
\end{equation}
\begin{equation}
\mbox{ACC} = 1 - \frac{1}{N_t}\sum_{i = 1}^{N_t}|f_i-y_i|
\label{eq:acc}
\end{equation}
where $f_i$ is the $i^{th}$ prediction of the prediction vector $f$; $y_i$ is the corresponding ground-truth in the ground-truth vector $y$; $N_t$ is the number of clips; $\mathrm{Cov}$ is the covariance while $\sigma_f$ and $\sigma_y$ are the standard deviations of $f$ and $y$, respectively.

\subsection{Comparison to existing approaches}
\label{subsec:sota}

\noindent Table \ref{tb:sota_noxi} and Table \ref{tb:sota_vhq} compare the variations of the proposed models to the existing state-of-the-art audio-visual personality analysis approaches (automatic personality perception (APP) solutions) on the NoXi and the VHQ datasets. Table \ref{tb:pvalue_sota} presents the results achieved by our best systems (the graph representations of multi-modal CNNs that are learned using adaptive loss, independent parameter settings and the graph representations are constructed using end-to-end vertex and edge feature learning strategy) and the results achieved by other methods under both interaction scenarios with highlighted statistical significance. It can be observed that for both datasets, the predictions produced by the graph representations of the explored multi-modal CNNs are positively correlated with all self-reported personality traits. Specifically, these graph representations achieved PCC$>0.37$ for \textit{Con}, \textit{Ext}, and \textit{Neu} traits on the NoXI dataset, which shows significant advantages over the other listed methods. Meanwhile, the graph representations of the explored multi-modal model (A-MModal (M)) also achieved the best average ACC result and the second best PCC result in recognizing the self-reported personality traits under human-computer interaction scenarios, i.e., it generated the best PCC results between predictions and ground-truth of the \textit{Neu} trait with PCC of $0.363$, showing more than $8\%$ relative improvements over the second best method \cite{song2021self}.

It also can be seen from Table \ref{tb:reaction_noxi} that the explored person-specific CNNs can accurately predict the corresponding target subjects' facial reactions, which are evidenced by the promising PCC results (more than $0.75$ for all systems) and RMSE results of the facial reaction (the landmarks) generation. This indicates that the simulated person-specific cognition can reproduce similar facial reactions for the majority frames of the target subject's video. In other words, the explored person-specific CNN can accurately represent for the target subject's cognition over time (a video's duration). On the contrary, these person-specific CNNs have poor performance in reproducing other subjects' facial reactions with the PCC between the reproduced reactions landmarks and the ground-truth less than $0.1$, which means each person-specific CNN can not reflect other subjects' cognition for facial reactions (i.e., the simulated cognition of the target subject is different from others' cognition). As a result, we assume the proposed approach can partially encode the target subject's person-specific cognitive process that is stable over time but different from other subjects.


\textbf{Discussion:} In summary, the results presented above indicate that despite CNNs and humans having different cognitive mechanisms, if a CNN can simulate a subject's cognitive process for generating facial reactions, this CNN's architectural parameters are positively associated with the subject's self-reported personality traits. Compared to existing solutions that directly predict personality traits from non-verbal behaviours, the proposed approach that recognises self-reported personality traits from the simulated cognition seems a more reliable solution. It should be noted that the advantage of our approach in human-machine interaction scenarios is not as clear as the human-human interaction scenarios.

The proposed approach achieved the best PCC and ACC performance for all five traits in human-human interaction scenarios, with results on all traits being significantly better over other compared methods, but it only outperformed most APP solutions in human-machine scenarios. This can be explained by the fact that the virtual human used in the VHQ dataset only has limited non-verbal facial behaviours, they are not as rich as real human speakers in the NoXI dataset. Thus, the listeners' facial reactions in human-machine interaction scenarios may be less correlated with the non-verbal behaviours expressed by the virtual human, resulting in the explored person-specific CNNs not being able to accurately represent the listeners' cognitive processes. This is also reflected by the worse facial reaction prediction performance generated for the VHQ dataset (Table  \ref{tb:reaction_vhq}). We also observed that the approaches which predict personality using video-level features (e.g., CR-Net, PALs, Spectral and the proposed approach) have clear advantages over the approaches that infer personality from a single frame or a thin slice (DCC and NJU-LAMDA), demonstrating that long-term information is more reliable for modelling self-reported personality traits. Finally, it is clear that the performance in recognising \textit{Neu} and \textit{Ext} traits are better than the other traits, which is consistent with what has been frequently claimed by previous studies \cite{kumari2004personality,kernberg2016personality}, i.e., \textit{Ext} and \textit{Neu} traits are well associated with human cognition.

\setlength{\tabcolsep}{2pt}
\begin{table}[t!]
	\begin{center}
\resizebox{1\linewidth}{!} {
		\begin{tabular}{|l| c| c c c c c l|}
			\toprule
 &Methods & Ope  & Con  & Ext &  Agr & Neu & Avg.   \\
\hline \hline
\multirow{8}*{ACC}

& DCC  \cite{guccluturk2016deep}  & 0.755 & 0.787 & 0.772 & 0.736 & 0.791 & 0.768 \\

& NJU-LAMDA  \cite{wei2018deep} & 0.741 & 0.826 & 0.827 & 0.753 & 0.789 & 0.787  \\

& Spectral \cite{song2020spectral}  & 0.868 & 0.909 & 0.903 & 0.898 & 0.910 & 0.898\\

& CR-Net \cite{li2020cr} & 0.892  & 0.916 & 0.924 & 0.888 & 0.913 & 0.907  \\

& PALs  \cite{song2021self}  & 0.845 & 0.819 & 0.916 & 0.837 & 0.911 & 0.866 \\

& Ours (A-MModal (S))  & 0.871	& 0.911	& 0.915	&0.903	&0.916	&0.903  \\

& Ours (MModal (M))  & \textbf{0.902} & 0.919  & 0.927	& \textbf{0.923}	&0.926	&0.919  \\

& Ours (A-MModal (M))   & 0.895 &	\textbf{0.925} &	\textbf{0.928} &	0.920 & \textbf{0.931} & \textbf{0.920} \\

\hline

\multirow{8}*{PCC}

& DCC \cite{guccluturk2016deep}  & -0.153 & -0.078 & 0.037 & -0.024 & 0.121 & 0.008 \\

& NJU-LAMDA  \cite{wei2018deep} & -0.110 & 0.118 & 0.115 & -0.067 & 0.032 & 0.017  \\

& Spectral \cite{song2020spectral}  & 0.135 & 0.246 & 0.265 & 0.192 & 0.277 &0.223 \\

& CR-Net  \cite{li2020cr} & 0.181 & 0.271 & 0.301 & 0.177 & 0.325 & 0.251  \\

& PALs \cite{song2021self}  & 0.129 & 0.091 & 0.270 & 0.106 & 0.264 & 0.172 \\

& Ours (A-MModal (S)) & 0.161&	0.322&	0.333&	0.239&	0.358&	0.283	\\

& Ours (MModal (M))  & \textbf{0.196}&	0.354&	0.403&	0.281&	0.450&	0.337   \\

& Ours (A-MModal (M))   & 0.189 & \textbf{0.376}	& \textbf{0.420}	& \textbf{0.289}	 &\textbf{0.481} &\textbf{0.351} \\

			\bottomrule
		\end{tabular}
        }
	\end{center}
	\caption{Personality recognition results on the NoXi dataset. \emph{MModal} denotes the graph representations of the explored multi-modal (audio-visual) CNNs. \emph{(M)} and \emph{(S)} represent the multi-level and single-level fusion, respectively. \emph{A-} represents that the CNNs were trained with adaptive loss. All employed CNNs are explored with the same depth (block distillation setting) as this setting achieved the best performance. For all our systems, the graph representations were obtained using OP-LW (VEN) vertices features and end-to-end learned edge features while GatedGCN was employed as the personality recognition model.} 
\label{tb:sota_noxi}
\end{table}
\setlength{\tabcolsep}{1.4pt}

\setlength{\tabcolsep}{2pt}
\begin{table}[t!]
	\begin{center}
\resizebox{1\linewidth}{!} {
		\begin{tabular}{|l| c| c c c c c l|}
			\toprule
 &Methods & Ope  & Con  & Ext &  Agr & Neu & Avg.   \\
\hline \hline
\multirow{8}*{ACC}

& DCC  \cite{guccluturk2016deep}  & 0.835 &0.837 & 0.840 & 0.838 & 0.840 & 0.838 \\

& NJU-LAMDA  \cite{wei2018deep} & 0.842 &	0.838 & 0.839 & 0.841 &	0.841 & 0.840  \\

& Spectral \cite{song2020spectral}  & 0.838 &0.842 & 0.843 & 0.846 & 0.846 & 0.843 \\

& CR-Net \cite{li2020cr} & \textbf{0.845} &	0.842 & 0.840 & 0.844 &	0.845 & 0.843  \\

& PALs  \cite{song2021self}  & 0.843 &	\textbf{0.843} &	0.843 & 0.846 & 0.845 & \textbf{0.844} \\

& Ours (A-SModal (S))  & 0.839	& 0.839	& \textbf{0.844}	&  \textbf{0.848}	&	0.846   & 0.843  \\

& Ours (MModal (M))  & 0.842 & 0.838  &  0.841  &  0.847   &   0.845 	& 0.843  \\

& Ours (A-MModal (M))  & 0.840 & 0.842 & 0.842 & 0.847 & \textbf{0.848} & \textbf{0.844} \\

\hline

\multirow{8}*{PCC}

& DCC \cite{guccluturk2016deep}  & -0.020 & 0.098 & 0.133 & -0.072 & 0.185 & 0.065 \\

& NJU-LAMDA \cite{wei2018deep} & 0.077 & 0.155 & 0.118 & 0.112 & 0.234 & 0.139  \\

& Spectral \cite{song2020spectral}  & -0.039 & 0.167 & 0.221 & 0.158 & 0.256 & 0.153 \\

& CR-Net  \cite{li2020cr} & \textbf{0.138} & \textbf{0.191} & 0.166 & 0.143 & 0.280 & 0.184  \\

& PALs \cite{song2021self}  & 0.135 & 0.187 & \textbf{0.279} & 0.132 & 0.336 & \textbf{0.214} \\

& Ours (A-SModal) & -0.021 & 0.127	& 0.267	& \textbf{0.193	}& 0.319	& 0.177 \\

& Ours (MModal)  & 0.088 & 0.136  & 0.192  &  0.191  & 0.358  &	0.193  \\

& Ours (A-MModal)   & 0.063 &	0.172 &	0.211 &	0.189 & \textbf{0.363} & 0.200 \\

			\bottomrule
		\end{tabular}
        }
	\end{center}
	\caption{Personality recognition results on the VHQ dataset. \emph{MModal} denotes the graph representations built on the face and sentence categorical features extracted from the speaker (please check Sec. \ref{subsec:imple}). The rest of the settings are the same as in Table \ref{tb:sota_noxi}.} 
\label{tb:sota_vhq}
\end{table}
\setlength{\tabcolsep}{2pt}

\setlength{\tabcolsep}{2pt}
\begin{table}[t!]
	\begin{center}
\resizebox{1\linewidth}{!} {
		\begin{tabular}{|l| c| c c c c c l|}
			\toprule
 &Traits & Ope  & Con  & Ext &  Agr & Neu & Avg.   \\
\hline \hline
\multirow{5}*{NoXI} 

&Spectrum (BP)\cite{song2020spectral} & \large{\textbf{+}}\footnotesize{(***)} & \large{\textbf{+}}\footnotesize{(***)} & \large{\textbf{+}}\footnotesize{(***)}  & \large{\textbf{+}}\footnotesize{(***)} & \large{\textbf{+}}\footnotesize{(***)} & \large{\textbf{+}}\footnotesize{(***)}  \\
&DCC$^{*}$ \cite{wei2018deep} & \large{\textbf{+}}\footnotesize{(***)}  & \large{\textbf{+}}\footnotesize{(***)}  & \large{\textbf{+}}\footnotesize{(***)}  & \large{\textbf{+}}\footnotesize{(***)}  & \large{\textbf{+}}\footnotesize{(***)} & \large{\textbf{+}}\footnotesize{(***)}  \\
&NJU-LAMDA$^{*}$ \cite{guccluturk2016deep}  &\large{\textbf{+}}\footnotesize{(***)}  & \large{\textbf{+}}\footnotesize{(***)}  & \large{\textbf{+}}\footnotesize{(***)}  & \large{\textbf{+}}\footnotesize{(***)}  & \large{\textbf{+}}\footnotesize{(***)} & \large{\textbf{+}}\footnotesize{(***)}  \\
&CR-Net$^{*}$ \cite{guccluturk2016deep}  &\large{\textbf{+}}\footnotesize{(***)}  & \large{\textbf{+}}\footnotesize{(***)}  & \large{\textbf{+}}\footnotesize{(***)}  & \large{\textbf{+}}\footnotesize{(***)}  & \large{\textbf{+}}\footnotesize{(***)} & \large{\textbf{+}}\footnotesize{(***)}  \\
&PALs$^{*}$ \cite{guccluturk2016deep}  &\large{\textbf{+}}\footnotesize{(***)}  & \large{\textbf{+}}\footnotesize{(***)}  & \large{\textbf{+}}\footnotesize{(***)}  & \large{\textbf{+}}\footnotesize{(***)}  & \large{\textbf{+}}\footnotesize{(***)} & \large{\textbf{+}}\footnotesize{(***)}  \\

\hline

\multirow{5}*{VHQ} 

&Spectrum (BP)\cite{song2020spectral} & \large{\textbf{+}}\footnotesize{(***)} & \large{\textbf{-}} & \large{\textbf{-}}  & \large{\textbf{+}}\footnotesize{(**)} & \large{\textbf{+}}\footnotesize{(***)} & \large{\textbf{+}}\footnotesize{(***)}  \\
&DCC \cite{wei2018deep} & \large{\textbf{+}}\footnotesize{(***)}  & \large{\textbf{+}}\footnotesize{(***)}  & \large{\textbf{+}}\footnotesize{(***)}  & \large{\textbf{+}}\footnotesize{(***)}  & \large{\textbf{+}}\footnotesize{(***)} & \large{\textbf{+}}\footnotesize{(***)}  \\
&NJU-LAMDA \cite{guccluturk2016deep}  &\large{\textbf{+}}\footnotesize{(***)}  & \large{\textbf{+}}\footnotesize{(*)}  & \large{\textbf{+}}\footnotesize{(***)}  & \large{\textbf{+}}\footnotesize{(***)}  & \large{\textbf{+}}\footnotesize{(***)} & \large{\textbf{+}}\footnotesize{(***)}  \\
&CR-Net \cite{li2020cr}  &\large{\textbf{+}}\footnotesize{(***)}  & \large{\textbf{+}}\footnotesize{(*)}  & \large{\textbf{+}}\footnotesize{(***)}  & \large{\textbf{+}}\footnotesize{(***)}  & \large{\textbf{+}}\footnotesize{(***)} & \large{\textbf{+}}\footnotesize{(***)}  \\
&PALs \cite{song2020spectral}  &\large{\textbf{+}}\footnotesize{(***)}  & \large{\textbf{-}}  & \large{\textbf{+}}\footnotesize{(***)}  & \large{\textbf{+}}\footnotesize{(***)}  & \large{\textbf{+}}\footnotesize{(*)} & \quad \large{\textbf{-}}  \\

			\bottomrule
		\end{tabular}
        }
	\end{center}
	\caption{Statistical significance testing results in terms of PCC achieved by our best system and the five reproduced systems on the NoXI and the VHQ dataset, where  \textbf{+} / \textbf{-} denotes that there is / there is no statistically significant difference between our approach and the other approach (The significance level of $*$ $P < 0.05$, $**$ $P < 0.01$, $***$ $P < 0.001$). To conduct the T-Test, we used the 7-fold results on the NoXI dataset. For VHQ dataset, we conducted 10 times leave-one-subject-out cross-validation for all approaches and the used these 10 results to compute the P-values.}
\label{tb:pvalue_sota}
\end{table}

\begin{table}[t]
\begin{center}
\resizebox{0.76\linewidth}{!}{ 
\begin{tabular}{|  c | c| c|}
\hline
Cognitive model &  PCC  & MSE ($\times 10^{-5}$)  \\
\hline
Audio-to-face & 0.769 & \textbf{2.882}  \\
Face-to-face & \textbf{0.770} &  2.894  \\
\hline
PS-Multi-to-face (S) & \textbf{0.781} &  2.390\\
IP-Multi-to-face (S) & 0.775 & 2.503 \\
A-IP-Multi-to-face (S) & \textbf{0.781} & \textbf{2.260}\\
\hline
PS-Multi-to-face (M) &  0.794 & 2.362 \\
IP-Multi-to-face (M) & 0.798 & \textbf{2.245} \\
A-IP-Multi-to-face (M) & \textbf{0.802} & 2.331 \\
A-IP-Dep-Multi-to-face (M) & 0.649  & 7.190  \\
\hline
\end{tabular}
}
\end{center}
\caption{Facial reactions prediction results on the NoXI dataset. \emph{PS-} and \emph{IP-} denote the parameter sharing and independent parameter strategy, respectively; \emph{Multi-} refers to the multi-modal audio and face features of the speaker were used as the input; \emph{A-} represents that the CNNs were trained with adaptive loss; \emph{Dep-} denotes that the depth is considered as a variable during the CNN search.}
\label{tb:reaction_noxi}
\end{table}

\begin{table}[t]
\begin{center}
\resizebox{0.76\linewidth}{!}{ 
\begin{tabular}{|  c | c| c|}
\hline
Cognitive model &  PCC  & MSE ($\times 10^{-5}$)  \\
\hline
Face-to-face & 0.596 & 6.632   \\
PS-Multi-to-face  & 0.588 & 6.550 \\
IP-Multi-to-face  & 0.602 & 6.279 \\
A-IP-Multi-to-face & 0.612 & \textbf{6.098} \\
A-IP-Dep-Multi-to-face & \textbf{0.619} & 6.177  \\
\hline
\end{tabular}
}
\end{center}
\caption{Facial reactions prediction results on the VHQ dataset. The \emph{Multi-} in this table refers to the face and sentence categorical features of the speaker were used as the input.}
\label{tb:reaction_vhq}
\end{table}

\subsection{Influence of demographics}
\label{subsec:interactive}

\begin{figure*}
	\centering
	\subfigure[Personality recognition results on the NoXI dataset.]{\label{subfig:noxi_inter}
		\includegraphics[width=15.6cm]{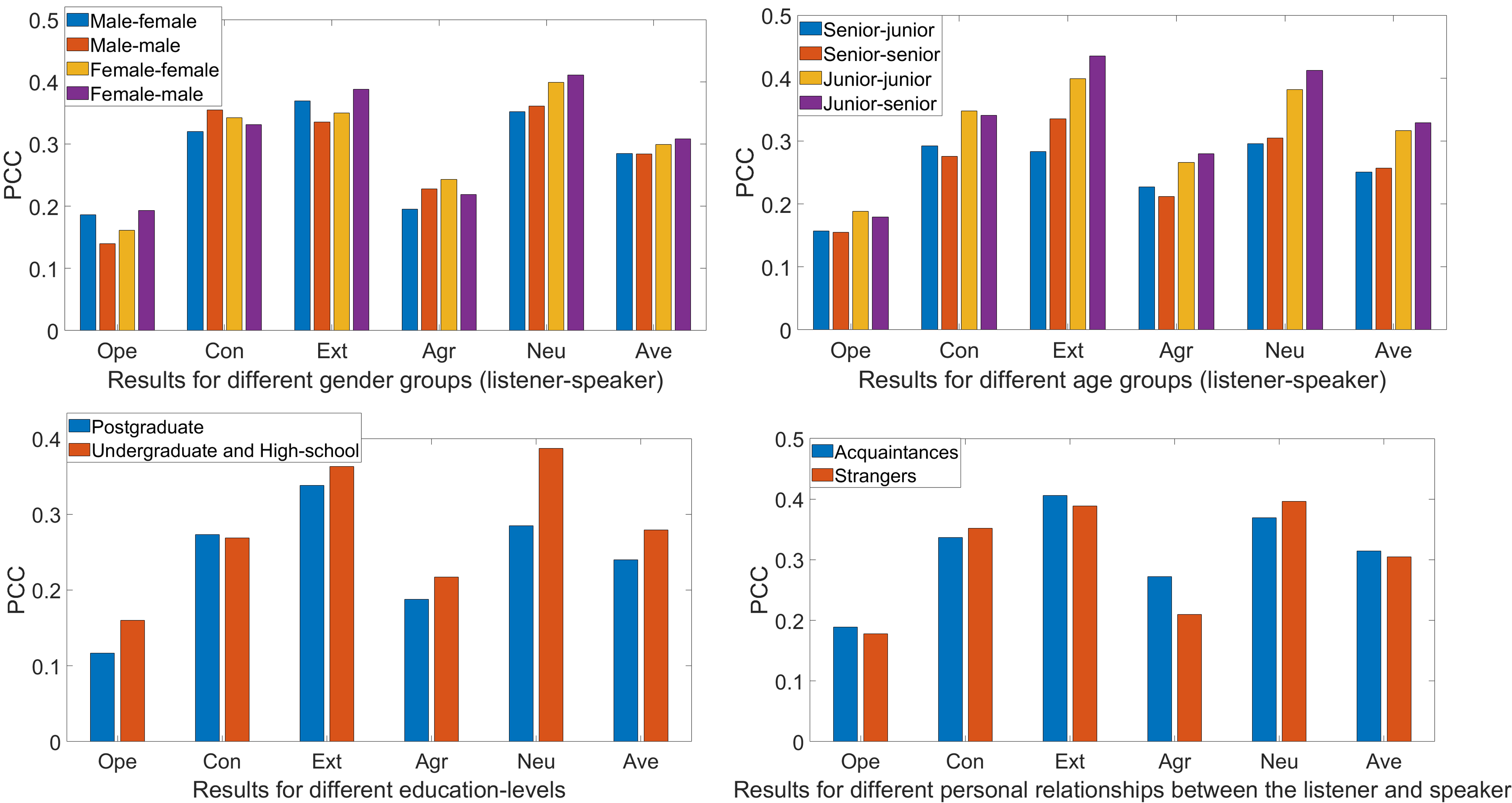}}
	\subfigure[Personality recognition results on the VHQ dataset.]{\label{subfig:VHQ_inter}
		\includegraphics[width=15.6cm]{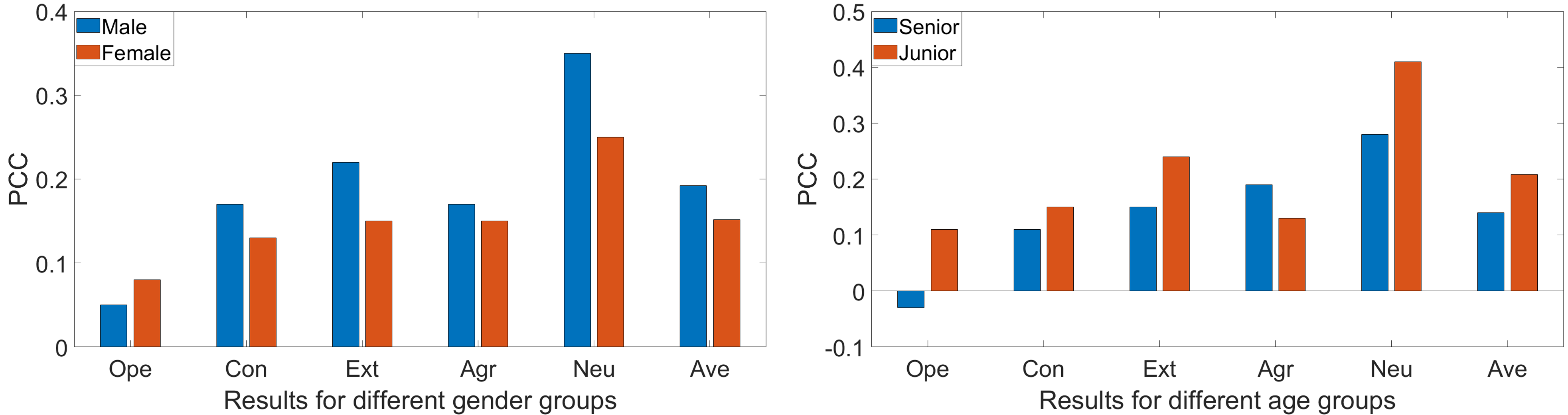}}
	\caption{The results of demographic studies. } 
    \label{fig:inter_setting_results}
\end{figure*}

\noindent This section investigates the performance variations of the proposed approach for different demographic groups. In particular, we investigate the influence of subjects' age, gender, education level as well as the personal relationship between the speaker and the listener in human-human dyadic interaction setting (Fig. \ref{subfig:noxi_inter}). We also investigate the influence of subjects' age and gender in human-machine dyadic interaction setting (VHQ dataset) (Fig. \ref{subfig:VHQ_inter}). Due to the limited number of data for each demographic group, we conducted subject-independent leave-one-out cross-validation for all interactive behaviour experiments. All results reported in this section are achieved by the graph representations of A-MModal(M), where end-to-end vertex and edge features were employed, and GatedGCN was used as the personality recognition model. The statistical significance testing results achieved by different demographic groups are listed in Table. \ref{tb:pvalue_interactive}.

\textbf{Gender:} We first investigate the difference in the results across different gender attributes. The NoXI dataset contains $62$ males, $26$ females and a participant with unidentified gender. There are $30$ males, $24$ females and a participant with unidentified gender in the VHQ dataset. 

In human-human dyadic interaction, it can be seen that the cognition simulated for female listeners' facial reactions is generally more informative than that of males for inferring personality, where the female listeners' person-specific CNNs explored during female-to-male dyadic interaction are the best predictors for \textit{Ope}, \textit{Ext} and \textit{Neu} traits among all gender groups. These results suggest that male listeners may be less willing to express their true personalities (emotions/ideas) through their facial reactions during human-human interactions while females tend to show their true personalities when having interactions with males. However, it can be seen that when males interact with a female, the simulated cognitive processes of their facial reactions is a relative good predictor for the \textit{Ope} trait, which can be explained by the fact that males usually more active when interacting with a female than interacting with another male \cite{mulac1988male,mulac1989men}. Meanwhile, the results achieved in the human-machine interaction scenario show that males' facial reactions generally reveal more personality cues, which contradicts the result achieved in human-human interaction setting. We hypothesize that this is because males usually have more interests in new technical applications \cite{goswami2015gender,barbieri2020gender} (e.g., virtual human \cite{jaiswal2019automatic,jaiswal2019virtual}), and thus may express more active behaviours that are informative for personality than females when interacting with the virtual human. Importantly, it can be seen from Table. \ref{tb:pvalue_interactive} that there are significant differences between the results achieved for male listeners and female listeners on both datasets, indicating that gender is a significant factor that influences the performance of the proposed approach.



\textbf{Age:} We then evaluate the impact of the age differences on personality recognition performance in both interaction scenarios. For the NoXI dataset, participants were categorized into $6$ age groups: 18-20 (7 participants), 21-25 (37 participants), 26-30 (30 participants), 31-35 (9 participants), 36-40 (3 participants), and 41-45 (3 participants). For the VHQ dataset, $28$ participants were younger than $26$ years old ($20.39$ years old in average) while the rest were $26$ years old or above ($30.96$ years old in average). For both datasets, we categorize participants whose age is younger than $26$ as the \textbf{junior} group, and the rest as the \textbf{senior} group. This categorization strategy aims to keep both age groups having similar number of subjects, preventing the recognition results from being largely influenced by unbalanced training data.

As illustrated in Fig. \ref{subfig:noxi_inter}, the explored person-specific CNNs of \textbf{junior subjects} are more reliable for representing their true personality as the yielded average PCC performance is $27.27\%$ superior than the senior listeners. This suggests that junior subjects' facial reactions are more likely to be expressed based on their true personality during the interaction, especially when they interact with senior subjects. In contrast, when a senior listener is interacting with a junior speaker, the senior listeners' facial reactions are less representative of their true personality. It can be observed that even in the human-machine interaction setting, the person-specific CNNs learned from junior subjects' facial reactions show clear advantages over those learned from senior participants in inferring true personality traits. According to Table \ref{tb:pvalue_interactive}, the results achieved across junior listeners and senior listeners have significant differences for all traits on both datasets. These results may indicate that the proposed approach is sensitive to subjects' age. In particular, junior subjects are more likely to express their true personalities through their facial reactions during a dyadic interaction. The potential reasons for these results is that junior individuals' facial expressions are usually more indicative of their real emotions \cite{folster2014facial}. In other words, senior individuals' facial behaviours may not directly reflect their state of mind and personalities.


\textbf{Education level:} Table. \ref{tb:pvalue_interactive} also shows that the education level has an impact on the recognition of \textit{Ope}, \textit{Ext}, \textit{Agr} and \textit{Neu} traits, where in our experiments $40$ subjects belong to either High-school level ($8$ subjects) or Undergraduate level ($33$ subjects) and $49$ subjects have Postgraduate level ($33$ subjects have Master degree and $16$ subjects are Doctorate level). As we can see, the persons-specific CNNs of subjects with higher education levels are less reliable for recognising most of their personality traits, e.g., \textit{Ope}, \textit{Ext}, \textit{Agr} and \textit{Neu}. The education level brought a large difference in Neu trait recognition, where the PCC result achieved for the Undergraduate and High-school group has $35.8\%$ relative improvement over the Postgraduate group. These results may suggest that facial reactions of subjects with higher education levels are less informative for their true personality. This contradicts the conclusion of \cite{kirouac1985accuracy} that facial expressions of subject with higher education are more informative for their emotion. However, we found that our results may be caused by the large differences in subjects' age, i.e. the age distribution of postgraduate level subjects are: $0$ (below 20 years old), $35$ (21-30 years old), $12$ (31-40 years old) and $2$ (above 40 years old) while the age distribution of undergraduate and high-school level subjects are: $7$ (below 20 years old), $32$ (21-30 years old), $0$ (31-40 years old) and $1$ (above 40 years old).

\textbf{Personal relationship:} We finally evaluate how the personal relationship in a dyadic interaction would affect the proposed approach. In the NoXI dataset, there are $47$ sessions conducted between acquaintances ($26$ pairs of acquaintances, $15$ pairs of friends and $6$ pairs of very good friends) and $37$ sessions conducted between strangers. As we can see from Table. \ref{tb:pvalue_interactive}, the relationship between two subjects also has a significant impact on the recognition of \textit{Ext}, \textit{Agr}, and \textit{Neu} traits. When acquaintances interact with each other, the explored person-specific CNNs can better reflect the \textit{Con}, \textit{Ext}, and \textit{Neu} traits. But if the subjects are strangers, the simulated cognitive processes from listeners' facial reactions are more informative for inferring the Agr and Ope traits. However, the differences caused by the subjects' relationships are generally not as significant as age, gender and education level (Table \ref{tb:pvalue_interactive}).


\setlength{\tabcolsep}{2pt}
\begin{table}[t!]
	\begin{center}
\resizebox{1\linewidth}{!} {
		\begin{tabular}{|l| c| c c c c c c |l|}
			\toprule
 &Traits & Ope  & Con  & Ext &  Agr & Neu & Avg.   \\
\hline \hline
\multirow{4}*{NoXI} 

&Gender  &\large{\textbf{+}}\footnotesize{(***)}  & \large{\textbf{+}}\footnotesize{(***)}  & \large{\textbf{+}}\footnotesize{(***)}  & \large{\textbf{+}}\footnotesize{(***)}  & \large{\textbf{+}}\footnotesize{(***)}  & \large{\textbf{+}}\footnotesize{(***)}\\
&Age &\large{\textbf{+}}\footnotesize{(***)}  & \large{\textbf{+}}\footnotesize{(***)}  & \large{\textbf{+}}\footnotesize{(***)}  & \large{\textbf{+}}\footnotesize{(***)}  & \large{\textbf{+}}\footnotesize{(***)} & \large{\textbf{+}}\footnotesize{(***)} \\
&Education  &\large{\textbf{+}}\footnotesize{(***)}  & \large{\textbf{-}}  & \large{\textbf{+}}\footnotesize{(***)}  & \large{\textbf{+}}\footnotesize{(***)}  & \large{\textbf{+}}\footnotesize{(***)} & \large{\textbf{+}}\footnotesize{(***)} \\
&Acquisition  &\large{\textbf{-}}  & \large{\textbf{-}}  & \large{\textbf{+}}\footnotesize{(***)}  & \large{\textbf{+}}\footnotesize{(***)}  & \large{\textbf{+}}\footnotesize{(***)} & \large{\textbf{+}}\footnotesize{(***)} \\

\hline

\multirow{2}*{VHQ} 

&Gender  &\large{\textbf{+}}\footnotesize{(***)}  & \large{\textbf{+}}\footnotesize{(***)}  & \large{\textbf{+}}\footnotesize{(***)}  & \large{\textbf{+}}\footnotesize{(***)}  & \large{\textbf{+}}\footnotesize{(***)} & \large{\textbf{+}}\footnotesize{(***)} \\
&Age &\large{\textbf{+}}\footnotesize{(***)}  & \large{\textbf{+}}\footnotesize{(***)}  & \large{\textbf{+}}\footnotesize{(***)}  & \large{\textbf{+}}\footnotesize{(***)}  & \large{\textbf{+}}\footnotesize{(***)} & \large{\textbf{+}}\footnotesize{(***)} \\

			\bottomrule
		\end{tabular}
        }
	\end{center}
	\caption{Statistical significance testing results in terms of the PCC achieved by different demographic groups, where  \textbf{+} / \textbf{-} denotes that there is / there is no statistical significant difference between our approach (The significance level of $*$ $P < 0.05$, $**$ $P < 0.01$, $***$ $P < 0.001$).}
\label{tb:pvalue_interactive}
\end{table}

\subsection{Ablation studies}
\label{subsec:ablation}

\noindent In this section, we explicitly investigate the sensitivity of the proposed approach to different person-specific CNN search settings (i.e., modality, parameter sharing strategy, loss function and topology alignment) and graph representation settings (i.e., vertex feature and edge feature). 

\subsubsection{Person-specific CNN settings}

\begin{figure*}
	\centering
	\subfigure[Personality recognition results on the NoXI dataset.]{\label{subfig:nas_noxi_bar}
		\includegraphics[width=15.6cm]{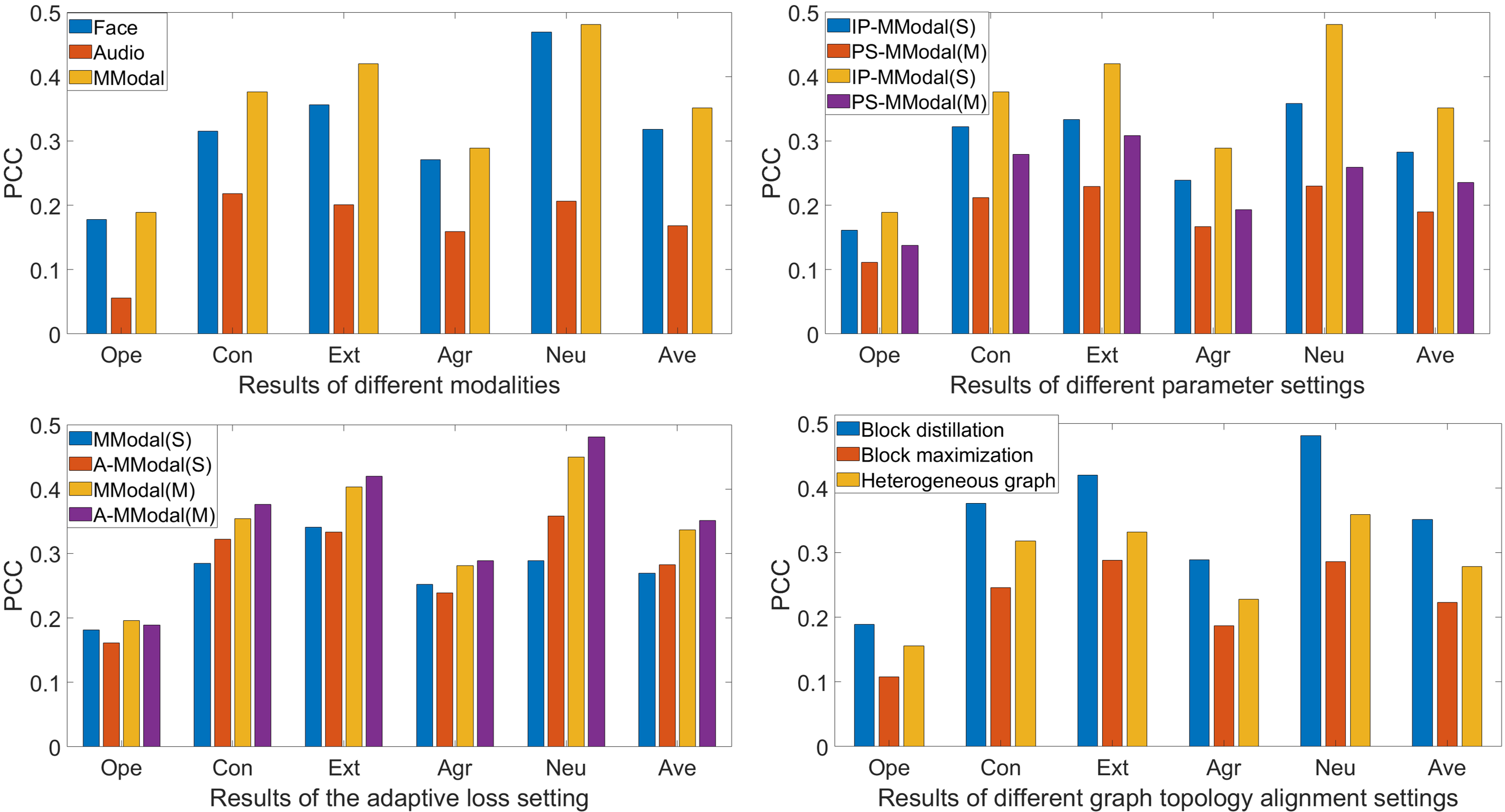}}
	\subfigure[Personality recognition results on the VHQ dataset.]{\label{subfig:nas_vhq_bar}
		\includegraphics[width=15.6cm]{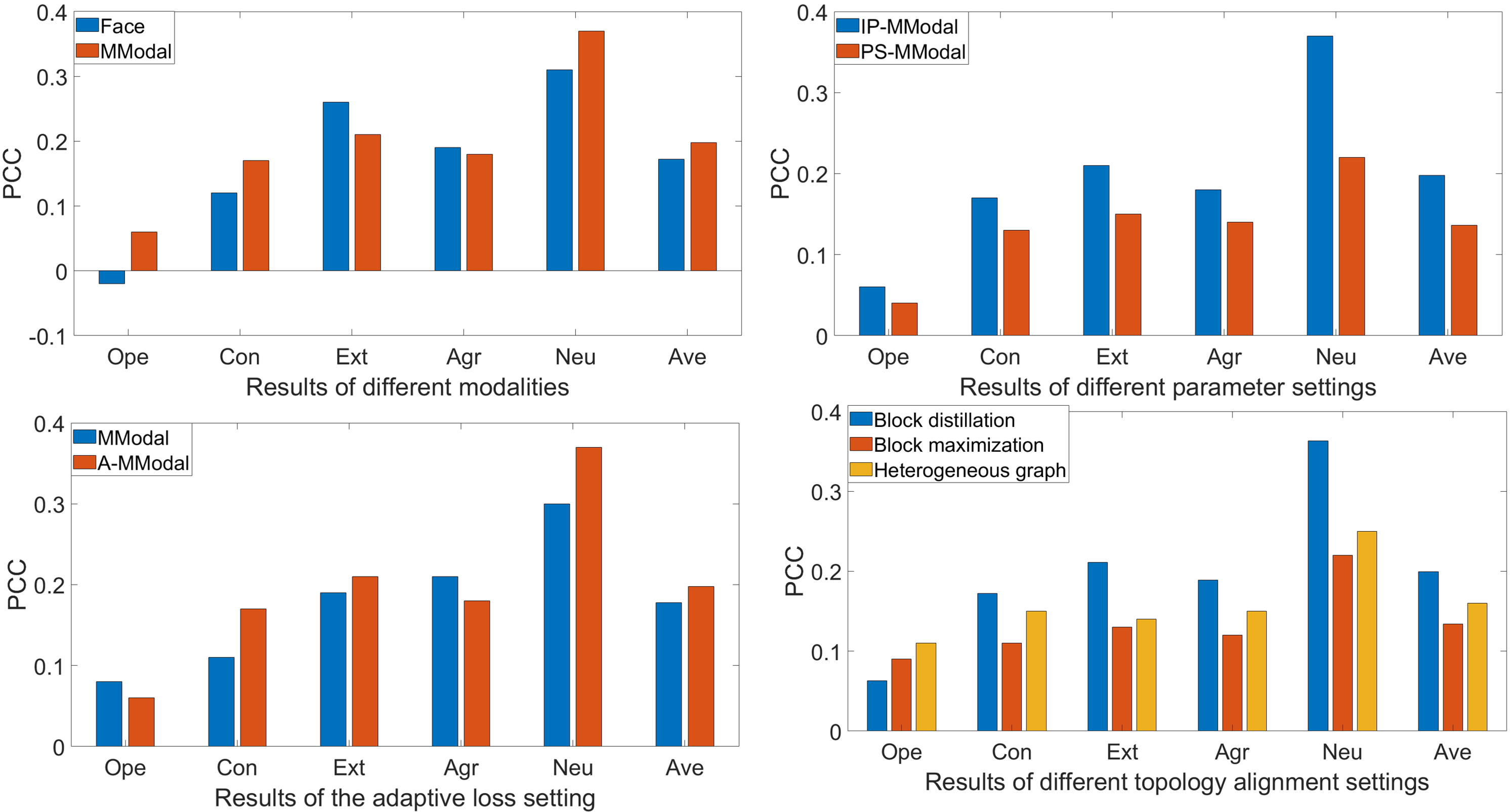}}
	\caption{The results of different person-specific CNN settings. The definition of \emph{MModal}, \emph{S}, \emph{M}, \emph{PS-}, \emph{IP-}, \emph{A-} can be found in the captions of Table. \ref{tb:sota_noxi}, Table. \ref{tb:sota_vhq}, Table. \ref{tb:reaction_noxi} and Table. \ref{tb:reaction_vhq}.} 
    \label{fig:nas_setting_results}
\end{figure*}

\noindent We first show the personality recognition performance achieved by different person-specific CNN settings in Fig. \ref{subfig:nas_noxi_bar} and Fig. \ref{subfig:nas_vhq_bar}. For all experiments, if not specifically mentioned, the default settings for CNN topology, parameter sharing strategy, loss function and topology alignment are: multi-modal, independent parameters (IP), adaptive loss, and block distillation, respectively. The statistical significance testing results achieved by the default setting and the second best system of each setting are shown in Table. \ref{tb:pvalue_interactive}.

\textbf{Modalities:} Firstly, it can be observed from the results on the NoXI dataset that predictions generated by all settings are positively correlated with the self-reported values across all five traits. In general, the multi-modal system achieved better results than single-modal systems for both personality recognition and facial reaction generation tasks. This demonstrates that both non-verbal audio and facial behaviours of the speaker contribute to the listener's facial reactions, where each of them contain some unique aspects in forming reactions. In other words, the cognitive processes triggered by each modality provides unique and useful clues for personality recognition, except for the audio not providing a significant contribution in helping the multi-modal model for recognizing the Neu trait (Table. \ref{tb:pvalue_nas}).

Meanwhile, as we can see from the results on the VHQ dataset, even simply adding the encoded virtual human sentence categorical feature (explained in Sec. \ref{subsec:imple}) can improve the recognition performance of Ope, Con and Neu traits. In comparison to the real human speaker, the facial behaviours of the virtual human may not be the key factor to trigger the listener's facial reactions, and thus the explored person-specific CNNs may not learn good hypotheses of the listeners' cognitive processes. However, the sentence categorical feature provides the context information which provides a controlled condition for a subject's reaction as well as strong supervision for the search of person-specific CNNs. Thus, adding sentence categorical feature as the extra modality improves the recognition of most traits.

\textbf{Parameter sharing strategies:} For the results achieved on both datasets, it is clear that graph representations of persons-specific CNNs searched by the independent parameter (IP) strategy have significant advantages over the results achieved by the widely-used parameter sharing (PS) strategy \cite{liu2018darts,pham2018efficient} over all five traits. These results validate our assumption that human cognition consists of a set of cognitive processes, each of which can be different from others, and thus each part of the explored CNN should have its own weights to better simulate a unique cognitive process/function.

\textbf{Loss function settings:} As we can see from the results on the NoXI dataset, despite most of our systems trained with standard MSE loss already achieved good performance in recognising Con, Ext and Neu traits, using the proposed adaptive loss provided further improvements. Meanwhile, the system that used the adaptive loss achieved the similar results in recognising Ope and Agr traits with no significant differences. Specifically, the use of adaptive loss still brought more than $5.8\%$ average improvement for Con, Ext and Neu traits. It can be observed from the results on human-machine interactions, the graph representations of multi-modal CNNs trained with adaptive loss better recognised all five traits. We also found that the systems that used the adaptive loss generated better facial reaction results. Since Neu and Ext traits can be better reflected by human cognitive processes \cite{kumari2004personality,kernberg2016personality}, we hypothesize that the proposed adaptive loss can partially address the uncertainty of subjects'  responding time, allowing the explored CNNs to better simulate target subjects' cognitive processes.

\textbf{Depth and topology alignment settings:} As we can see from Fig. \ref{subfig:nas_noxi_bar} and Fig. \ref{subfig:nas_vhq_bar}, directly feeding the heterogeneous graph representations of the explored person-specific CNNs to the GatedGCN generate a better recognition result than aligning them to isomorphic graph representations using the block maximization alignment, which means the block maximization setting may largely distort the original information carried by person-specific CNNs. More importantly, under both interaction scenarios, isomorphic graph representations achieved by the block distillation showed advantages over all five traits compared to other settings. In this paper, the block distillation is achieved by setting all blocks of all person-specific CNNs having the same depth. Moreover, it can be found from Table \ref{tb:reaction_noxi} and Table \ref{tb:reaction_vhq} that the person-specific CNNs with their unique depth do not show clear advantages in reproducing listeners' facial reactions. These results suggest that there is no clear improvement of the depth-independent setting on facial reaction generation, which means even CNNs that were searched using the same depth have a comparable or even better capability to represent the target subjects' cognitive processes. Meanwhile, the typologies of the produced heterogeneous graph representations can varied a lot, which may lead the training process of the corresponding GCNs to become more difficult. Consequently, the isomorphic graph representations aligned by the block distillation outperformed the heterogeneous graph representations for personality recognition with significant improvements. In addition, applying the block maximization setting to encode the variable-size person-specific CNNs into the isomorphic graph representation may result in the graph representations being very sparse for CNNs whose blocks do not have the full depth, and using the identity mapping to represent the redundant cells in the block may not accurately reflect the exact cognitive process of corresponding subjects, which can be a key issue that contributes to the relatively poor recognition performance.

\setlength{\tabcolsep}{2pt}
\begin{table}[t!]
	\begin{center}
\resizebox{1\linewidth}{!} {
		\begin{tabular}{|l| c| c c c c c c |l|}
			\toprule
 &Traits & Ope  & Con  & Ext &  Agr & Neu & Avg.   \\
\hline \hline
\multirow{4}*{NoXI} 

&Face  &\large{\textbf{+}}\footnotesize{(**)}  & \large{\textbf{+}}\footnotesize{(***)}  & \large{\textbf{+}}\footnotesize{(*)}  & \large{\textbf{+}}\footnotesize{(*)}  & \large{\textbf{-}}  & \large{\textbf{+}}\footnotesize{(***)}\\
&PS &\large{\textbf{+}}\footnotesize{(***)}  & \large{\textbf{+}}\footnotesize{(***)}  & \large{\textbf{+}}\footnotesize{(***)}  & \large{\textbf{+}}\footnotesize{(***)}  & \large{\textbf{+}}\footnotesize{(***)} & \large{\textbf{+}}\footnotesize{(***)} \\
&N-Ap  &\large{\textbf{-}}  & \large{\textbf{+}}\footnotesize{(*)}  & \large{\textbf{+}}\footnotesize{(*)}  & \large{\textbf{-}} & \large{\textbf{+}}\footnotesize{(**)} & \large{\textbf{+}}\footnotesize{(***)} \\

&HGG  &\large{\textbf{+}}\footnotesize{(***)}  & \large{\textbf{-}} & \large{\textbf{+}}\footnotesize{(***)}  & \large{\textbf{+}}\footnotesize{(*)}  & \large{\textbf{+}}\footnotesize{(***)} & \large{\textbf{+}}\footnotesize{(***)} \\

\hline

\multirow{4}*{VHQ} 

&Face  &\large{\textbf{+}}\footnotesize{(***)}  & \large{\textbf{+}}\footnotesize{(***)}  & \large{\textbf{+}}\footnotesize{(***)}  & \large{\textbf{-}}  & \large{\textbf{+}}\footnotesize{(**)} & \large{\textbf{+}}\footnotesize{(**)} \\
&PS &\large{\textbf{+}}\footnotesize{(***)}  & \large{\textbf{+}}\footnotesize{(***)}  & \large{\textbf{+}}\footnotesize{(***)}  & \large{\textbf{+}}\footnotesize{(***)}  & \large{\textbf{+}}\footnotesize{(***)} & \large{\textbf{+}}\footnotesize{(***)} \\
&N-Ap  &\large{\textbf{+}}\footnotesize{(***)}  & \large{\textbf{+}}\footnotesize{(***)}  & \large{\textbf{+}}\footnotesize{(*)}  & \large{\textbf{+}}\footnotesize{(**)}  & \large{\textbf{+}}\footnotesize{(***)} & \large{\textbf{+}}\footnotesize{(***)} \\

&HGG  &\large{\textbf{+}}\footnotesize{(**)}  & \large{\textbf{-}}  & \large{\textbf{+}}\footnotesize{(***)}  & \large{\textbf{+}}\footnotesize{(**)}  & \large{\textbf{+}}\footnotesize{(***)} & \large{\textbf{+}}\footnotesize{(***)} \\

			\bottomrule
		\end{tabular}
        }
	\end{center}
	\caption{Statistical significance testing results difference in terms of PCC achieved by our default system (using multi-modal, independent parameters (IP), adaptive loss, and block distillation) and the second best systems of each person-specific CNN setting (modality, weights sharing, loss function and graph topology alignment (HGG denotes the heterogeneous graph)) on NoXI and VHQ dataset, where  \textbf{+} / \textbf{-} denotes that there is / there is no statistical significant difference between our approach (The significance level of $*$ $P < 0.05$, $**$ $P < 0.01$, $***$ $P < 0.001$)..}
\label{tb:pvalue_nas}
\end{table}

\subsubsection{Graph representation}
\label{subsec:results_graph}

\begin{figure*}
	\centering
	\subfigure[Personality recognition results on the NoXI dataset.]{\label{subfig:graph_noxi_bar}
		\includegraphics[width=16cm]{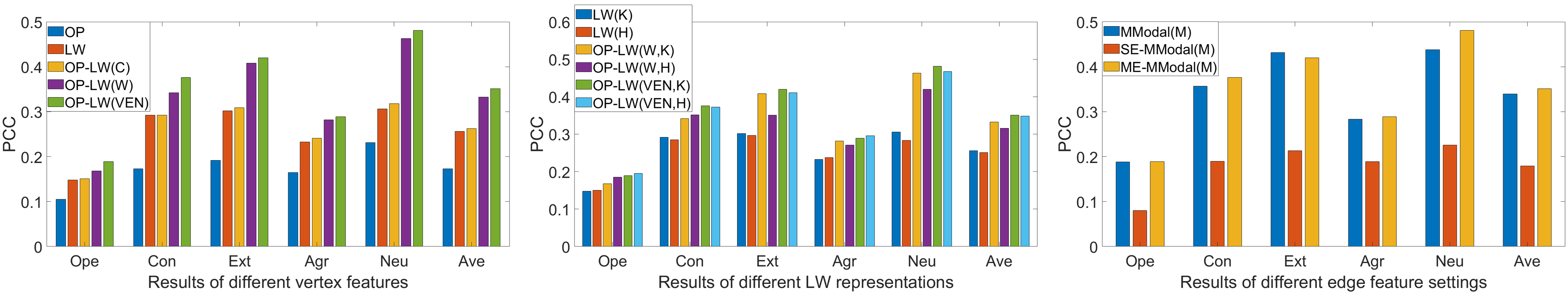}}
	\subfigure[Personality recognition results on the VHQ dataset.]{\label{subfig:graph_vhq_bar}
		\includegraphics[width=16cm]{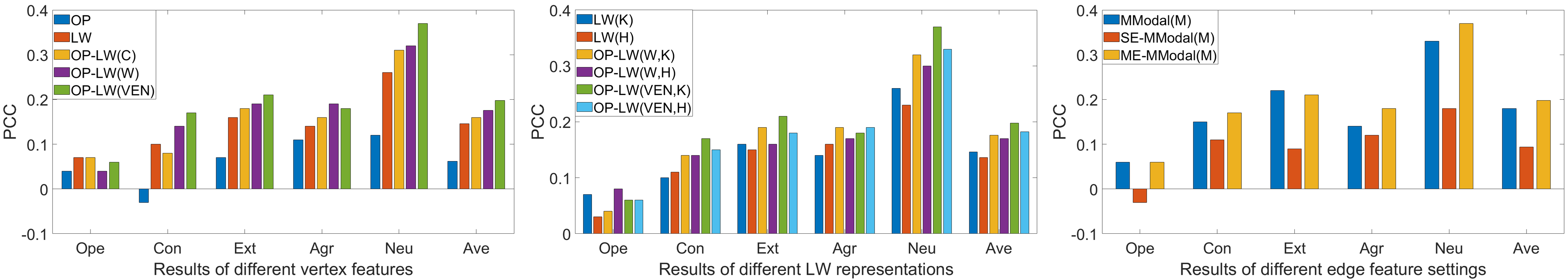}}
	\caption{The results of different graph representation and personality recognition model settings. The definition of different vertex settings can be found in Sec. \ref{subsec:results_graph}; The \emph{SE} and \emph{ME} represent the results achieved by using a single shared ERN and a set of ERNs to deep learn multi-dimensional edge features, respectively.} 
    \label{fig:graph_setting_results}
\end{figure*}

\noindent In this section, we demonstrate the advantages of the proposed end-to-end vertex feature and edge feature learning strategy for constructing graph representations in Fig. \ref{fig:graph_setting_results}. For all experiments, if not specifically mentioned, the default settings for vertex feature, LWs representation and edge feature are: OP-LW (VEN) vertex feature, LWs representation that contains of the top-5 kernels' weights, and edge features learned by multiple ERNs. The statistical significance testing results achieved by our best graph representation setting and the second best settings are listed in Table \ref{tb:pvalue_graph}.

\textbf{Vertex feature settings:} We first compare the proposed deep-learned vertex feature to four hand-crafted vertex features. It can be seen that the graph representations of the LW and OP-LW (C) vertex features have a similar capability for recognising personality, both of which outperformed the graph representations that only use OPs as the vertex feature. This can be explained by the fact that the OP vertex feature ignores all LWs which are crucial in deciding CNNs' generalization capabilities, i.e., the clues for inferring personality traits reside in both OPs and their LWs. TMeanwhile, the OP-LW (C) feature does not show a clear advantage over the LWs feature, whose performance is also not comparable to the OP-LW (W) and OP-LW (VEN) features, demonstrating that simply concatenating OPs and LWs is not a proper way to combine their clues. In other words, the best recognition results of all five traits are achieved either by OP-LW (VEN) feature or OP-LW (W) vertex feature.  As a result, we concluded that using each OP to weight corresponding LWs is a more superior way to combine them. Moreover, the OP-LW (VEN) setting shows significant advantages over the OP-LW (W) on Ope and Con traits under human-human interaction and Ope, Con. Ext, and Neu traits under human-machine interaction. Thus, we assume that the OP-LW (VEN) setting allows a better weighting vector to be learned to construct the each vertex, which not only considers the original OPs but also task-specific information. In addition, the figure also compares the results achieved by different LWs representations. While the results achieved by two LWs representations varyies under different settings, they have very similar results for most traits, i.e., there is no significant differences between the results achieved by two LWs settings over all five traits inunder the human-human interaction setting as well as three traits under the human-machine interaction, where using weights of Top-5 kernels performed slightly better than the use of histogram. In short, the clues provided by both LWs representations are similarly informative for the recognition of most personality traits' recognition.


\textbf{Edge feature settings:} We also compare the proposed end-to-end learned multi-dimensional edge features to the widely-used binary adjacency edge feature (0 or 1). It can be seen that the multi-dimensional edge features deep-learned using multiple ERNs outperformed the results achieved by the graph representations that only use binary adjacency matrix to define the relationship between vertices, with more than $3.4\%$ and $10\%$ average improvements under human-human and human-machine interaction scenarios, respectively, i.e., the improvements brought by these edge features are significant for most traits (Con, Ext, Neu in the human-human interaction setting and Con, Agr, Neu in the human-machine interaction setting) as well as the average performance. Such results validate the usefulness of the proposed end-to-end multi-dimensional edge feature learning strategy, which can better describe the relationship between adjacent vertices with task-specific clues, particularly for the Con and Neu trait. Meanwhile, we also found that using a single ERN to define all edge features generated very poor personality recognition performance, which means this strategy failed to learn the task-specific relationship between different pairs of vertices. This can be explained by the fact that each pair of vertices has their unique relationship which can be largely different from others, and thus a single ERN may not be able to jointly learn all of them.

\subsubsection{Personality recognition model}

\begin{figure}
	\centering
	\subfigure[Results on the NoXI dataset.]{\label{subfig:backend_noxi}
		\includegraphics[width=8cm]{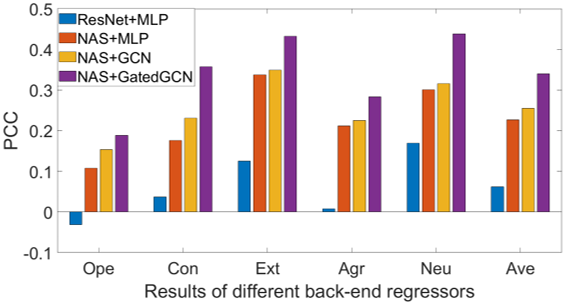}}
	\subfigure[Results on the VHQ dataset.]{\label{subfig:backend_vhq}
		\includegraphics[width=8cm]{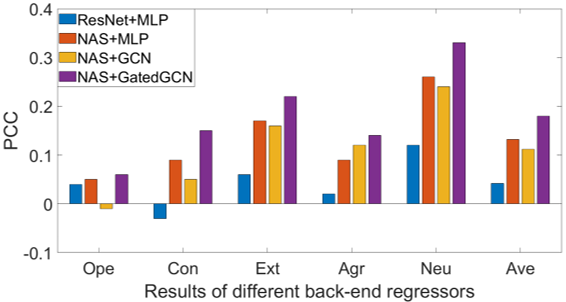}}
	\caption{The results achieved by different back-end models.} 
    \label{fig:backend}
\end{figure}

\noindent Fig. \ref{fig:backend} compares the influence of different back-end models on learning the produced graph representations for personality recognition. Firstly, it is important to note that predictions of all three models achieved positive correlations across all traits under both interaction scenarios but the GCN model produced the negative result in predicting the Con trait under the human-machine interaction. These results demonstrate that the explored person-specific CNN architectures are indeed positively associated with target subjects' personality traits. Importantly, it is clear that in comparison to the standard GCN, the GatedGCN is clearly a better model to project the proposed graph representation to the target subjects' self-reported personality traits. Also, the statistical significant difference between the PCC results achieved by our graph representation (processed by GatedGCN) and the concatenation of person-specific CNN parameters (processed by MLP) are listed in Table. \ref{tb:pvalue_graph}, showing that the use of the proposed graph representation yielded significantly better results.

It should be emphasized that even the system that simply uses an MLP with four hidden layers to process person-specific representations (i.e., 1-D vectors that are produced by concatenating all OP and LW of each person-specific processor, respectively and then processed by correlation-based feature selection (CFS) \cite{hall1999correlation}) can make predictions that show clear positive correlations ($PCC > 0.1$) with the self-reported Con, Ext, Agr and Neu traits. In particular, the setting of the MLP system treats each directed edge as a standard CNN block that contains $10$ parallel branches, and the output of the block is the weighted sum of all branches, i.e. OPs play the role of LWs in this setting. Thus, the provided feature vector can also be treated the same as the feature vector that concatenates all weights of a standard CNN. In this sense, we also train a person-specific ResNet-50 network for each person based on the proposed adaptive loss, where each network's weights are concatenated as a 1-D vector which is then processed by CFS. Then, we follow the same setting as 'NAS+MLP' to use such vectors for personality recognition. It can be observed from Fig. \ref{fig:backend} that the vectors came from the explored person-specific CNNs generated significant better results than these of ResNet-50 networks, which shows that the CNNs explored by the proposed approach can better simulate person-specific cognitive processes. This is also evidenced by the better facial reaction generation performance displayed in Table \ref{tb:reaction_noxi} and Table \ref{tb:reaction_vhq}.

\setlength{\tabcolsep}{2pt}
\begin{table}[t!]
	\begin{center}
\resizebox{1\linewidth}{!} {
		\begin{tabular}{|l| c| c c c c c c |l|}
			\toprule
 &Traits & Ope  & Con  & Ext &  Agr & Neu & Avg.   \\
\hline \hline
\multirow{4}*{NoXI} 

&OP-LW(W) &\large{\textbf{+}}\footnotesize{(**)}  & \large{\textbf{+}}\footnotesize{(**)}  & \large{\textbf{-}}  & \large{\textbf{-}}  & \large{\textbf{-}}  & \large{\textbf{+}}\footnotesize{(*)}\\
&OP-LW(VEN,H) &\large{\textbf{-}}  & \large{\textbf{-}}  & \large{\textbf{-}} & \large{\textbf{-}}  & \large{\textbf{-}} & \large{\textbf{-}} \\
&NE-MModal &\large{\textbf{-}}  & \large{\textbf{+}}\footnotesize{(**)}  & \large{\textbf{+}}\footnotesize{(*)}  & \large{\textbf{-}}  & \large{\textbf{+}}\footnotesize{(***)} & \large{\textbf{+}}\footnotesize{(**)} \\
&GCN &\large{\textbf{+}}\footnotesize{(***)}  & \large{\textbf{+}}\footnotesize{(***)}  & \large{\textbf{+}}\footnotesize{(***)}  & \large{\textbf{+}}\footnotesize{(***)}  & \large{\textbf{+}}\footnotesize{(***)} & \large{\textbf{+}}\footnotesize{(***)} \\

\hline

\multirow{4}*{VHQ}

&OP-LW(W)  &\large{\textbf{+}}\footnotesize{(***)}  & \large{\textbf{+}}\footnotesize{(***)}  & \large{\textbf{+}}\footnotesize{(***)}  & \large{\textbf{-}}  & \large{\textbf{+}}\footnotesize{(**)} & \large{\textbf{+}}\footnotesize{(***)} \\
&OP-LW(VEN,H) &\large{\textbf{-}} & \large{\textbf{-}}  & \large{\textbf{+}}\footnotesize{(**)}  & \large{\textbf{-}}  & \large{\textbf{+}}\footnotesize{(*)} & \large{\textbf{+}}\footnotesize{(**)} \\
&NE-MModal  &\large{\textbf{-}} & \large{\textbf{+}}\footnotesize{(*)}  & \large{\textbf{-}} & \large{\textbf{+}}\footnotesize{(***)}  & \large{\textbf{+}}\footnotesize{(*)} & \large{\textbf{+}}\footnotesize{(**)} \\
&MLP &\large{\textbf{+}}\footnotesize{(***)}  & \large{\textbf{+}}\footnotesize{(***)}  & \large{\textbf{+}}\footnotesize{(***)}  & \large{\textbf{+}}\footnotesize{(***)}  & \large{\textbf{+}}\footnotesize{(***)} & \large{\textbf{+}}\footnotesize{(***)} \\

			\bottomrule
		\end{tabular}
        }
	\end{center}
	\caption{Statistical significance testing results in terms of PCC achieved by our default system (OP-LW (VEN) vertex feature, LWs representation that contains weights of top-5 kernels, edge features learned by multiple ERNs, and GatedGCN) and the second best model of different graph/back-end model settings (vertex feature, LWs representation, edge feature, and back-end model) on the NoXI and VHQ dataset. \textbf{+} / \textbf{-} denotes that there is / there is no statistical significant difference between our approach (The significance level of $*$ $P < 0.05$, $**$ $P < 0.01$, $***$ $P < 0.001$).}
\label{tb:pvalue_graph}
\end{table}

\section{Conclusions and future work}

\noindent This paper presents the first work which recognises true personality traits from the person-specific cognitive processes that are simulated from audio-visual non-verbal data. The person-specific cognitive process of a target subject is represented by an automatically searched person-specific CNN architecture that takes the speaker's non-verbal audio-visual and targets to reproduce the target subject's (listener's) facial reactions during the dyadic interaction. To use the explored CNN architecture as a representation for personality recognition, we proposed a novel graph encoding approach that deep learns a task-specific graph representation from the CNN's architectural parameters.

The experimental results achieved under human-human and human-machine dyadic interaction scenarios suggest the following conclusions: (i). the graph representations learned from person-specific CNNs are positively associated with the target subjects' self-reported personality traits, showing that the cognitive process of the explored CNNs also have the similar personality as their corresponding target subjects, i.e., the explored CNNs may have their own personalities; (ii). the proposed approach has clear advantages over most existing APP approaches which predict personality directly from non-verbal behaviours, demonstrating that it is reliable to recognise self-reported (true) personality from subjects' internal states; (iii). we found that the graph representations learned by the proposed approach are informative for recognising Ext and Neu traits under both interaction scenarios; (iV). the proposed approach performed better personality recognition and facial reaction prediction under the human-human interaction scenario than the human-machine scenario, indicating that nonverbal behaviours expressed by human speakers are not only more correlated with the speakers' reaction but also more powerful to trigger the listeners' personality-related facial behaviours; (V). many human attributes (e.g., age, gender, education level, and interpersonal relationship) can influence the performance of the proposed approach, where the gender and age are influential factors. This is caused by that the facial reactions of a similar intention or emotion can be varied due to the these factors; (vi). among several technical settings of the proposed approach, the proposed adaptive loss function, independent parameters strategy and end-to-end vertices/edges feature learning strategies have largely enhanced the personality recognition performance.


The main limitation of this work is that searching for a unique CNN architecture for each subject takes a relatively long period, i.e., the training and inference duration of the proposed approach are expected to be longer than most existing approaches. Therefore, it may not be suitable for \emph{fast} personality assessment requirements. Another limitation is that we only used audio-visual modalities but ignored other human signals such as psychological signals (EEG, heart rates, skin temperature, etc.) and verbal information, which contribute important information to one's communication and reactions. In addition, the employed datasets only contain relatively small number of videos, which may not enough to train a perfect, large CNN-GCN personality recognition model (contains a VEN, a set of ERNs and a GatedGCN).

As a result, a potential future direction would be to improve the person-specific cognitive process simulation algorithm so that it does not require searching for a person-specific CNN for each person from scratch, allowing the faster person-specific graph representation generation. Then, additional modalities (e.g., verbal signal, psychological signals, etc.) might enable the CNNs to be more similar to the target subjects’ cognition in a dyadic interaction, where dialogue response generation can be utilised to predict listeners’ verbal responses. All these modalities can in principle be combined via the proposed fusion module, i.e., combining them at multiple levels, as each is influenced by the others. There remain of course some modality-specific issues to resolve, such as differences in sampling rates or noise levels, and more, so while it’s definitely possible there is also substantial future research to be done in this area. Meanwhile, from the application perspective, this work opens up a new avenue of research for predicting and recognizing socio-emotional phenomena (personality, affect, engagement, etc.) from the simulations of person-specific cognitive processes that will have further implications for relevant fields including neuroscience, and cognitive, behavioural and emotion sciences. Specifically, it can be applied to analyze mental health or other human internal states with domain-specific loss functions under clinical settings, i.e. representing them with CNN parameters, or used for creating data-driven robot coaches that can express personalized behaviours during dyadic interactions.

\backmatter


\section*{Acknowledgement}

\noindent Siyang Song and Hatice Gunes are partially supported by the European Union’s Horizon 2020 Research and Innovation Programme, under the WorkingAge Project (grant agreement No.826232).  Hatice Gunes is also supported by the EPSRC (Grant Ref: EP/R030782/1). Linlin Shen and Zilong Shao are supported by the National Natural Science Foundation of China under Grant 91959108.

\bibliographystyle{plain}
\bibliography{sn-bibliography}


\end{document}